\pdfoutput=1

\documentclass[11pt]{article}

\usepackage{acl}

\usepackage{times}
\usepackage{latexsym}

\usepackage[T1]{fontenc}

\usepackage[utf8]{inputenc}

\usepackage{microtype}

\usepackage{array}
\usepackage{amsmath}
\usepackage{bm}
\usepackage{graphicx}
\usepackage{enumitem}
\usepackage{lipsum}
\usepackage{cleveref}
\usepackage{booktabs}
\usepackage{tikz}
\usepackage{amssymb}
\usepackage{arydshln}
\usepackage{multirow}
\usepackage{xcolor}
\usepackage{color}
\usepackage{xspace}
\usepackage{nccmath}
\usepackage{pifont}
\newcommand{\cmark}{\ding{51}}%
\newcommand{\xmark}{\ding{55}}%
\setlist[itemize]{leftmargin=*}

\newcommand{\ie}{\emph{i.e.,}\xspace}
\newcommand{\eg}{\emph{e.g.,}\xspace}

\title{Parameter-Efficient Conversational Recommender System \\
as a Language Processing Task}
\author{Mathieu Ravaut$^{1}$, {\bf Hao Zhang}$^{1}$, {\bf Lu Xu}$^{2}$, {\bf Aixin Sun}$^{1}$, {\bf Yong Liu}$^{1}$ \\
        $^{1}$School of Computer Science and Engineering, Nanyang Technological University, Singapore \\
        $^{2}$Singapore University of Technology and Design\\
        \texttt{mathieuj001@e.ntu.edu.sg} \\
}

\begin{document}
\maketitle
\begin{abstract}
Conversational recommender systems (CRS) aim to recommend relevant items to users by eliciting user preference through natural language conversation. Prior work often utilizes external knowledge graphs for items’ semantic information, a language model for dialogue generation, and a recommendation module for ranking relevant items. This combination of multiple components suffers from a cumbersome training process, and leads to semantic misalignment issues between dialogue generation and item recommendation. In this paper, we represent items in natural language and formulate CRS as a natural language processing task. Accordingly, we leverage the power of pre-trained language models to encode items, understand user intent via conversation, perform item recommendation through semantic matching, and generate dialogues. As a unified model, our PECRS (Parameter-Efficient CRS), can be optimized in a single stage, without relying on non-textual metadata such as a knowledge graph. Experiments on two benchmark CRS datasets, ReDial and INSPIRED, demonstrate the effectiveness of PECRS on recommendation and conversation. Our code is available at: \url{https://github.com/Ravoxsg/efficient_unified_crs}.
\end{abstract}

\section{Introduction}
\label{sec:1}
Conversational recommender systems (CRS) have become an active research topic, which leverages both natural language processing and recommendation techniques to provide high-quality recommendations through interactive conversations with users~\cite{survey-jannach,survey-gao,survey-pramod}. 

\begin{figure}[t]
    \includegraphics[width=\columnwidth]{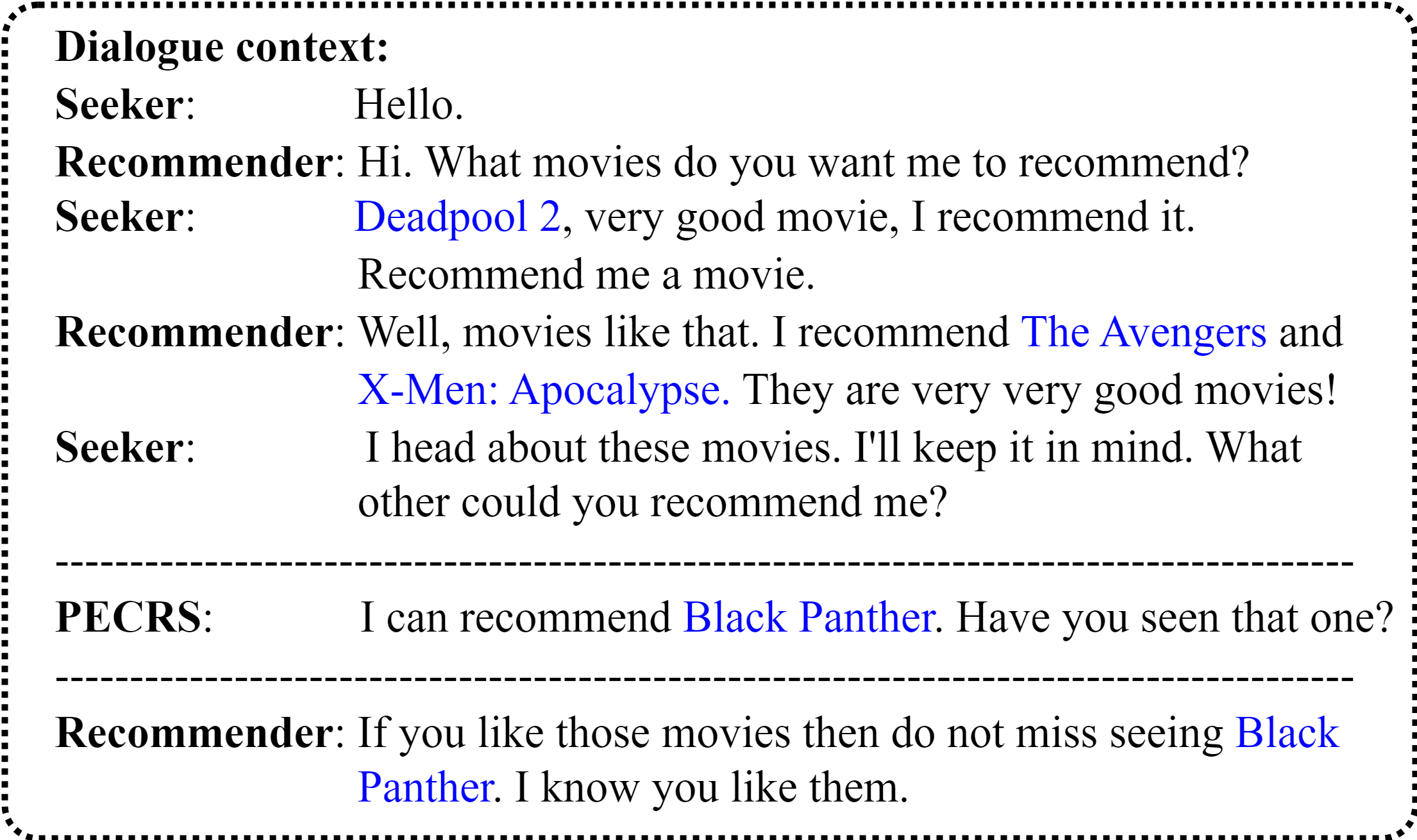}
    \caption{\small An example of dialogue from ReDial~\cite{redial}, where \textcolor{blue}{blue} color denotes the movie items.}
    \label{fig:1}
\end{figure}

CRS consists of two sub-tasks: 1) generating natural language responses to interact with user (\textit{conversation}); and 2) recommending desirable items to user based on dialogue context (\textit{recommendation}). An example of CRS data and model prediction is shown in \Cref{fig:1}. In general, CRS represents a significant advancement in the field of recommendation, which could be applied to various possible use cases, such as e-commerce, entertainment and content platforms.

Existing CRS methods can be roughly categorized into \emph{attribute-based} and \emph{generation-based} methods. The attribute-based methods~\cite{ear,crsal,qrec} focus on collecting user preferences on item attributes to narrow down recommendation space to items with desired properties. The generation-based methods~\cite{kgsf,c2crs,unicrs} aim to acquire feedback from users, generate natural responses, and establish a comprehensive understanding of conversation to recommend the most desirable items to user. In this work, we focus on generation-based CRS, which was greatly facilitated with the rise of task-specific CRS datasets like ReDial~\cite{redial}, INSPIRED~\cite{inspired}, TG-ReDial~\cite{tgredial} and DuRecDial~\cite{durecdial}.

The key challenge of CRS methods consists in how to jointly model language generation and item recommendation, which are tasks of entirely different natures. Early approaches~\cite{kbrd,kgsf,kecrs,c2crs} mainly model conversation and recommendation tasks separately by incorporating external knowledge graphs (KG) for item semantics and designing auxiliary strategies to enhance the interactions between two tasks. They generally treat items as nodes, which neglects the affluent textual information of items. They also sustain semantic misalignment issue due to inconsistent item and word representations, because conversation and recommendation modules are separately learned. Recent approaches~\cite{recindial,barcor,unicrs,mese} explore to seamlessly integrate conversation and recommendation modules for better knowledge sharing and semantic alignment via unified frameworks. However, due to the natural gap between recommendation and conversation, they still require multiple training phases~\cite{unicrs} and/or additional modules~\cite{recindial,mese} to integrate the two tasks, failing to reach desired level of integration.

With the rapid development of language models (LMs), LMs for recommendation has gained significant attention. Based on LMs, recent work~\cite{wu2023aso,lin2023howcr} also shows a growing correlation between recommendation and language tasks. Thus, instead of applying structured KGs, we stick to using item text descriptions together with dialogue contexts for CRS, which formulates the CRS directly as a natural language processing task. Specifically, we devise a \textbf{P}arameter-\textbf{E}fficient \textbf{C}onversational \textbf{R}ecommender \textbf{S}ystem (\textbf{PECRS}), which jointly solves recommendation and conversation by training a single model once, to bypass the shortcomings of prior work in CRS. PECRS only relies on a frozen pre-trained LM as backbone and employs a parameter-efficient plugin module to unify response generation and item recommendation in a simple yet flexible manner. Besides, we design a shared negative sampling strategy to sample negative items across subtasks and data points within the same mini-batch to boost both training efficiency and model performance. Moreover, thanks to the parameter-efficient plugin module, PECRS can easily scale up to larger LM backbones without significantly increasing training parameters.
In brief, our contributions are the following:

\begin{itemize}
\setlength{\itemsep}{1pt}
    \item To the best of our knowledge, this is the first work solving CRS by optimizing a single model in a single training phase and bypassing the need for either KGs or additional item encoders.
    
    \item We demonstrate how to jointly generate response and learn item representations using a single and frozen language model. Through parameter-efficient fine-tuning techniques, our method is with low computation cost, and can easily scale to larger backbones for higher performance.
    
    \item Experiments on two benchmark datasets, ReDial and INSPIRED, demonstrate the effectiveness of our proposed PECRS method, which is competitive with SOTA.
\end{itemize}

\section{Related Work}
\label{sec:2}
Existing conversational recommender systems (CRS) can be roughly categorized into attribute-based and generation-based CRS methods. The attribute-based CRS methods utilize predefined actions to interact with users and target on accomplishing the recommendation task with fewer turns~\cite{crs,crs2,ear,crsal,qrec,crif}. Our work belongs to the generation-based CRS, which focuses on developing natural language based approaches to make high-quality recommendation and generate human-like responses simultaneously~\cite{redial,inspired,tgredial,durecdial}. 

Generation-based CRS methods usually devise a recommendation module and a conversation module to implement item recommendation and response generation, respectively. \citet{redial} propose the first CRS dataset named ReDial, and solve it via encoder-decoder-based dialogue generator and autoencoder-based recommender. Subsequent work commonly adopts external resources to incorporate sufficient contextual information for better performance. Numerous works~\cite{kbrd,kgsf,crfr,crwalker,kecrs,ntrd,uccr,mgcg,vricr} use knowledge graphs (KG)~\cite{dbpedia,conceptnet} coupled with graph networks~\cite{rgcn} to enhance the items and user preference understanding by designing sophisticated semantic alignment strategies. RevCore~\cite{revcore} and C$^2$-CRS~\cite{c2crs} further incorporate movie reviews to enrich the contextual knowledge via cross-attention~\cite{revcore} and contrastive learning~\cite{c2crs}. Despite consecutive improvements, these works rely on different architectures for conversation and recommendation, making them difficult to be effectively integrated for end-to-end training and knowledge sharing. Consequently, they still suffer from a mismatch between conversation and recommendation modules as well as inferior efficiency.

To remedy the aforementioned issues, recent approaches explore to jointly learn both conversation and recommendation tasks by pre-trained LMs. UniCRS~\cite{unicrs} adopts the DialoGPT~\cite{dialogpt} for both conversation and recommendation by tuning soft prompts~\cite{softprompt} dedicated to each task. Nevertheless, UniCRS requires three rounds of optimization, \ie semantic fusion pre-training, conversation tuning, and recommendation tuning. UniMIND~\cite{unimind} follows the UniCRS paradigm with BART~\cite{bart} as the backbone, which unifies multi-goal CRS, \ie multi-tasks, using prompting strategy with multiple training stages. RecInDial~\cite{recindial} augments items into DialoGPT vocabulary and designs a pointer mechanism for dynamic word and item prediction to achieve single multi-tasking process. Similarly, BARCOR~\cite{barcor} utilizes BART to recommend items with encoder and generate responses with decoder concurrently. Instead of using KG, MESE~\cite{mese} encodes item representations using metadata and fuses them into dialogue for joint conversation and recommendation learning using GPT-2~\cite{gpt2} as the backbone. Although these methods attempt to integrate conversation and recommendation tasks for joint optimization, they rely on extra modules (\eg R-GCN~\cite{rgcn} and DistilBERT~\cite{distilbert}) for either item encoding or semantic fusion, and multi-round training stages. In contrast, our goal is to design a framework to unify the CRS training under a single model optimized in a single training stage.

Our work also employs parameter-efficient fine-tuning (PEFT) strategies. PEFT, including prompt tuning~\cite{softprompt}, Adapters~\cite{adapters}, and LoRA~\cite{lora}, is a series of techniques to adapt (large) LMs with fewer parameters and low computation costs to achieve same or even better performance comparing to the standard fine-tuning on downstream tasks. PEFT has shown great promise in various natural language~\cite{llamaadapter,qlora}, computer vision~\cite{he2022parameterefficient,chen2023vision}, and recommendation~\cite{Fu2023ExploringAT} tasks, but remains underexplored in CRS area. In this work, we aim to train CRS via PEFT plugins without touching the parameters of the backbone LM.

\section{Methodology}
\label{sec:3}
\begin{figure*}[t]
    \includegraphics[width=\textwidth]{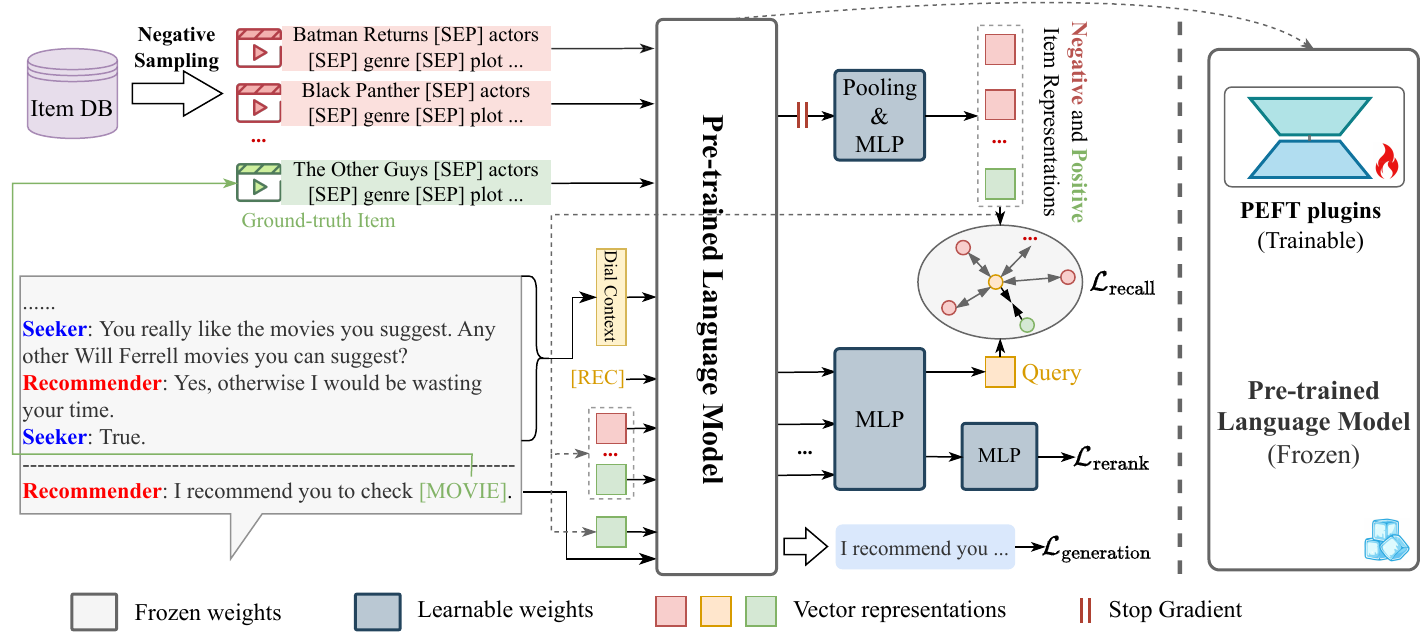}
    \caption{\small The overall architecture of the proposed Parameter-efficient Conversation Recommendation System (PECRS), where the PEFT denotes the parameter-efficient fine-tuning. Instead of fine-tuning backbone model, we inject PEFT plugins into backbone model and fine-tune the PEFT weights (see the figure in the right).}
    \label{fig:2}
\end{figure*}

In this section, we first describe the problem statement of conversational recommendation systems (CRS). Then we present the proposed \textbf{P}arameter-\textbf{E}fficient \textbf{C}onversational \textbf{R}ecommender \textbf{S}ystem (\textbf{PECRS}) method in detail. The overall architecture of PECRS is shown in \Cref{fig:2}.

\subsection{Problem Formulation}
\label{subsec:3.1}
Let $\bm{\mathcal{I}}=\{I_{1}, I_{2}, \ldots, I_{N_{\text{item}}}\}$ represent the item database, which contains $N_{\text{item}}$ unique items, and $\bm{\mathcal{D}}=\{D_{1}, D_{2}, \ldots, D_{N_{\text{dial}}}\}$ denote a CRS dataset with $N_{\text{dial}}$ dialogues. Each dialogue $D$ consists of $n_{\text{utt}}$ utterances denoted by $D=\{u_t\}_{t=1}^{n_{\text{utt}}}$, where $u_t$ represents the utterance at the $t$-th turn and each utterance $u_t=\{w_j\}_{j=1}^{n}$ contains a sequence of $n$ words. The task of CRS is to generate the response and recommend desirable items based on the given dialogue history and item database. To be specific, given the dialogue history up to the $t$-th turn $D_{t}=\{u_i\}_{i=1}^{t-1}$ and the item database $\bm{\mathcal{I}}$, the CRS needs to \textit{recommend} a set of candidate items $\bm{\mathcal{I}}_t$ from $\bm{\mathcal{I}}$, and \textit{generate} the response $u_t$ which includes the items $\bm{\mathcal{I}}_t$. The recommended candidate items set $\bm{\mathcal{I}}_t$ could be \textit{empty} when no recommendation is needed, or contain \textit{one} or \textit{more} items depending on the responses. 

In this work, we apply our method to the \emph{movie recommendation} (\ie $\bm{\mathcal{I}}$ denotes a movie items set), but the process would be identical with other types of items. We follow prior work~\cite{unicrs,mese} to adjust data samples and predict response with \emph{a single recommended movie per utterance}.

\subsection{Model Input}
\label{subsec:3.2}

In PECRS, items are represented by their textual descriptions, hence both input streams are modeled as text. Nevertheless, we design a few special tokens to distinguish the various elements in PECRS. 

\paragraph{Special Tokens.} 
Our PECRS is built upon a pre-trained LM under the decoder-only style, parameterized by $\theta$ (e.g, GPT-2). However, LMs generally do not have the capacity for recommendation task. Thus, we define four special tokens, \ie ``[ITEM]'', ``[SEP]'', ``[REC]'' and ``[REC\_END]'', and add them into the LM's vocabulary to guide the model's understanding of recommended items.

\paragraph{Item Metadata.}
Prior work~\cite{kgsf,kecrs,recindial,c2crs,unicrs} usually exploits external KG to encode item representations. They generally regard items as nodes and model relations among items through R-GCN~\cite{rgcn}, but neglect the rich textual descriptions of the items. In contrast, similar to \citet{mese}, we explore to use the static textual metadata of items. Item descriptions can be fed into a language model directly, hence bypassing the semantic misalignment issue. To be specific, each item $I_j$ is represented by affluent relevant information of the item rather than just its title. For movie recommendation, we use the following format ``\emph{Movie title [SEP] Actors [SEP] Director(s) [SEP] Genre(s) [SEP] Plot}'' to describe a movie item, where [SEP] is used to mark the separation among different fields. Note this process can be directly generalized to other domains by using the meta information of items in the target domain.
Formally, let $I_{j}=\{c_{j,k}\}_{k=1}^{l}$ denotes the $j$-th item textual data with $l$ tokens, its output from LM is $\bm{I}_j = [\bm{c}_{j,1}, \ldots, \bm{c}_{j,l}]$. We further adopt a MLP layer $h_{\text{item}}$ with learnable pooling weight $\bm{w}$ to aggregate the item representation as:

\begin{equation}
\label{eq:1}
    \bm{v}_{j} = h_{\text{item}}(\bm{w}^{T}\cdot\bm{I}_{j}).
\end{equation}

\paragraph{Dialogue Context.}
The dialogue context is made of all utterances up to the current $t$-th utterance: $D_{t}=\{u_i\}_{i=1}^{t-1}$. The word embeddings of the $i$-th utterance are denoted as $\bm{u}_i=[\bm{c}_{i,1},\dots,\bm{c}_{i,n}]$. If any utterance $u_{i}$ contains an item, it will be replaced by ``[ITEM]'' token and its item representation is also concatenated to the left side of the utterance's word embeddings. Otherwise, it remains unchanged. Let $\bm{v}_{\text{sep}}$, $\bm{v}_{\text{rec}}$ and $\bm{v}_{\text{rec\_end}}$ denote the token representations of ``[SEP]'', ``[REC]'' and ``[REC\_END]'', respectively. Suppose the $i$-th utterance contains an item, if it is from \emph{seeker}, its token embeddings are represented as $\tilde{\bm{u}}_i = [\bm{v}_{\text{sep}},\bm{v}_{j},\bm{v}_{\text{sep}},\bm{u}_i]$; if it is from \emph{recommender}, its token embeddings are $\tilde{\bm{u}}_i = [\bm{v}_{\text{rec}},\bm{v}_{j},\bm{v}_{\text{rec\_end}},\bm{u}_i]$. Thus, the token embedding sequences of dialogue context are the concatenation of all utterances with $\bm{v}_{\text{rec}}$ representation:

\begin{equation}
\label{eq:2}
    \bm{D}_t = [\bar{\bm{u}}_{1},\ldots,\bar{\bm{u}}_{t-1},\bm{v}_{\text{rec}}],
\end{equation}

where $\bar{\bm{u}}_i=\tilde{\bm{u}}_i$ if the utterance contains items, otherwise $\bar{\bm{u}}_i=\bm{u}_i$.

\subsection{Recommendation}
\label{subsec:3.3}

The recommendation module contains two processes: retrieval and re-ranking. The retrieval process is to select candidate items relevant to dialogue context from item database. The re-ranking process further re-ranks the selected candidate items after aggregating knowledge from the dialogue context.

\paragraph{Retrieval.}
We use the movie item in the response to be predicted as the ground-truth item, and sample $M$ negative items from item database. Then, we use their textual descriptions to encode item representations via \Cref{eq:1} and derive ground-truth item $\bm{v}_p$ and negative items $\{\bm{v}'_j\}_{j=1}^{M}$. 
As the dialogue context is ended with ``[REC]'' token (ref. \Cref{eq:2}) and decoder-only LM can aggregate all contextual information via causal self-attention, we utilize LM's output of ``[REC]'' token, denoted as $\bm{d}_{t}$, to represent \textit{query} representation of dialogue context. 
We adopt a noise-contrastive estimation (NCE)~\cite{nce,nce2,nce3} objective to bring together the query $\bm{d}_{t}$ with the positive key $\bm{v}_p$ and push apart $M$ negative (query, key) pairs formed by the set $\bm{\mathcal{N}} = \{(\bm{d}_t, \bm{v}'_j)\}_{j=1}^{M}$.

The NCE objective is written as:

\begin{equation}
\label{eq:3}
    \mathcal{E}_{D_t} = \frac{e^{f(\bm{d}_t)^{\top} \odot \bm{v}_p}}
    {e^{f(\bm{d}_t)^{\top} \odot \bm{v}_p} + \sum\limits_{(\bm{d}_t, \bm{v}'_j) \sim \bm{\mathcal{N}}} e^{f(\bm{d}_t)^{\top} \odot \bm{v}'_j}},
\end{equation}

where $f$ is a projection head with two-layer MLP and ReLU activation; $\odot$ denotes the \emph{angular} distance, $\sqrt{2(1-\cos(\bm{a},\bm{b}))}$, which measures the similarity between two vectors, $\bm{a}$ and $\bm{b}$. The recall loss for retrieval process is defined as:

\begin{equation}
\label{eq:4}
    \mathcal{L}_{\text{recall}} = - \frac{1}{|\bm{\mathcal{D}}|}\sum\limits_{D_t \in \bm{\mathcal{D}}} \log(\mathcal{E}_{D_t}).
\end{equation}

Note we stop the gradients of LM and only optimize the pooling and MLP layers for item representations encoding during training (ref. \Cref{fig:2}) to accelerate the learning process. The item representations will be reused in re-ranking process and the LM will be optimized at this stage accordingly. 

\paragraph{Re-ranking.}
The item representations derived from retrieval process are reused in the re-ranking process to aggregate the knowledge of dialogue context. To be specific, given both positive and negative items, we concatenate them with the token embeddings of dialogue context as $[\bm{D}_{t},\bm{v}_p, \bm{v}'_1,\ldots,\bm{v}'_M]$ and feed into LM then MLP $f$ to compute the context-aware item representations $[\bm{q}_p, \bm{q}_1,\ldots,\bm{q}_M]$. Note that we adopt a special attention mask to enforce that each item $\bm{v}_{j}$ only attends to tokens from $\bm{D}_{t}$, and positional embeddings are removed for item tokens to avoid any position leakage. Then another MLP layer $g$ is applied to compute the final item scores as $\bm{r}=[r_p,r_1,\ldots,r_M]$.

The training objective of re-ranking process is:

\begin{equation}
\label{eq:5}
    \mathcal{L}_{\text{rerank}} = \frac{1}{|\bm{\mathcal{D}}|} \sum\limits_{D_t \in \bm{\mathcal{D}}} f_{\text{XE}}(\bm{r}, \bm{Y}),
\end{equation}

where $\bm{Y}=[1,0,\dots,0]$ and $f_{\text{XE}}$ denotes cross-entropy loss. Note we shuffle $r$ and $\bm{Y}$ jointly to avoid the positional bias of ground-truth labels.
If a data point has no recommended item in the response, we set $ \mathcal{L}_{\text{recall}} = \mathcal{L}_{\text{rerank}} = 0 $.

\subsection{Response Generation}
\label{subsec:3.4}

The response generation aims to predict the current utterance $u_t=\{w_j\}_{j=1}^{n}$ by giving the dialogue context. During training, if the $u_t$ contains an item to be recommended, the representations of the ground-truth item is appended to the corresponding dialogue context to guarantee that the LM generates the response relevant to the item. Then, the input for response generation is:

\begin{equation}
\label{eq:6}
    \tilde{\bm{D}}_t = [\bar{\bm{u}}_1, \ldots, \bar{\bm{u}}_{t-1}, \bm{v}_{\text{rec}}, \bm{v}_p, \bm{v}_{\text{rec\_end}}].
\end{equation}

Otherwise, the input for response generation stays as $\tilde{\bm{D}_t} = [\bar{\bm{u}}_1, \ldots, \bar{\bm{u}}_{t-1}]$. In general, the response generation is optimized by the standard next-token prediction objective as:

\begin{equation}
\label{eq:7}
\resizebox{.98\hsize}{!}{
$\mathcal{L}_{\text{gen}} = - \frac{1}{|\bm{\mathcal{D}}|} \sum\limits_{D_t \in \bm{\mathcal{D}}} \frac{1}{n} \sum\limits_{j=1}^{n}\log(p_{\theta}(w_{j} | w_{1:(j-1)}, \tilde{\bm{D}}_t).$
}
\end{equation}

\subsection{Parameter-Efficient Learning}
\label{subsec:3.5}

We exploit parameter-efficient fine-tuning (PEFT) techniques for training. PEFT can achieve comparable performance to standard fine-tuning~\cite{llmadapter} with higher training efficiency and avoid the catastrophic forgetting issue of LM. Specifically, we leverage the LoRA~\cite{lora} method, which incorporates low-rank weight matrices into transformer layers to adapt LM to downstream tasks by fine-tuning the injected weights only. 
In addition to LoRA layers, we also fine-tune the task-specific MLP layers $f$, $g$ and $h_{\text{item}}$ and the token embeddings of the four special tokens. PECRS only updates a small proportion (around $5\%$) of the total number of parameters in the model.

\subsection{Training and Inference}
\label{subsec:3.6}

The PECRS is trained in a \emph{singe-stage} end-to-end manner by minimizing the following loss:

\begin{equation}
\label{eq:10}
    \mathcal{L} = \alpha\times\mathcal{L}_{\text{recall}} + \beta\times\mathcal{L}_{\text{rerank}} + \gamma\times\mathcal{L}_{\text{gen}},
\end{equation}

where $\alpha$, $\beta$ and $\gamma$ are hyperparameters to balance the three losses. During training, we randomly sample $M_{\text{train}}$ negative items and share them for computing the $\mathcal{L}_{\text{recall}}$ and $\mathcal{L}_{\text{rerank}}$ losses. Besides, we share the negative samples across batch elements and ensure that none of them is a positive for the dialogue contexts within a batch.

During inference, we first use PLM to encode the representations of all items in the database, which are reused for all dialogue contexts. Then the top-$M_{\text{inference}}$ items with highest similarities to the dialogue context query are retrieved via $f(\bm{d}_t)^{\top}\odot\bm{v}_j$ (see \Cref{eq:3}). We further re-rank the $M_{\text{inference}}$ items to obtain the top-1 item as the recommendation output. In practice, we set $M_{\text{train}} < M_{\text{inference}}$. We show that $M$ yields an important trade-off between efficiency and recommendation performance both during training and inference in \Cref{subsec:5.2}. 
Moreover, the predicted item is appended at the end of the dialogue context rather than the ground truth in \Cref{eq:6} in order to prompt the model for response generation. To determine whether a movie should be recommended at inference, we check whether the ``[ITEM]'' token is present in the generated response.

\section{Experiments}
\label{sec:4}
\subsection{Experimental Settings}
\label{subsec:4.1}

\paragraph{Datasets.} 
We conduct experiments on two commonly used datasets, \ie ReDial~\cite{redial} and INSPIRED~\cite{inspired}. ReDial\footnote{https://redialdata.github.io/website/} contains $11,348$ conversations ($10,006$ for train and $1,342$ for test) about movie recommendation between \emph{seeker} and \emph{recommender}, which is constructed through crowd-sourcing workers on Amazon Mechanical Turk. INSPIRED\footnote{https://github.com/sweetpeach/Inspired} is also about movie recommendation with smaller size of $999$ ($801$ for train, $99$ for development and $99$ for test) and more flexibility given to workers. The statistics of both datasets are summarized in \Cref{datasets}.

\begin{table}[]
\resizebox{\columnwidth}{!}{  
\fontsize{20pt}{20pt}\selectfont
\begin{tabular}{lllllll}

\toprule 

\textbf{Dataset} 
& \textbf{\begin{tabular}[c]{@{}l@{}}Unique\\ items\end{tabular}} 
& \textbf{Dialogues} 
& \textbf{Utterances} 
& \textbf{\begin{tabular}[c]{@{}l@{}}Recommender \\utterances\end{tabular}}
& \textbf{\begin{tabular}[c]{@{}l@{}}Rec. utt.\\ w/o rec.\end{tabular}} 
& \textbf{\begin{tabular}[c]{@{}l@{}}Rec. utt.\\ w/ rec.\end{tabular}} \\

\midrule 

ReDial      & 6,637 & 11,348 & 139,557 & 73,999 & 31,119 & 42,880 \\
INSPIRED    & 1,546 & 999 & 21,124 & 10,122 & 7,243 & 2,879 \\   

\bottomrule

\end{tabular}
}
\caption{\small Statistics on ReDial and INSPIRED datasets, combined over train, dev and test sets.}
\label{datasets}
\end{table}

\begin{table*}[]
\setlength{\tabcolsep}{5pt}
\resizebox{\textwidth}{!}{  
\fontsize{20pt}{20pt}\selectfont
\begin{tabular}{l c c c c c c c c c c c c c c}

\toprule 

\multirow{2}{*}{\textbf{Model}} 
& \multicolumn{3}{c}{\textbf{Metadata}}      
& \multicolumn{3}{c}{\textbf{Model Properties}}   
& \multicolumn{4}{c}{\textbf{ReDial}}                            
& \multicolumn{4}{c}{\textbf{INSPIRED}} \\


\cmidrule(lr){2-4}
\cmidrule(lr){5-7}
\cmidrule(lr){8-11}
\cmidrule(lr){12-15}

& \textbf{KG} 
& \textbf{Reviews}
& \textbf{Description}
& \textbf{Extra Model}
& \textbf{PEFT}
& \textbf{Rounds}
& \textbf{R@1} 
& \textbf{R@10} 
& \textbf{R@50} 
& \textbf{Unique}
& \textbf{R@1} 
& \textbf{R@10} 
& \textbf{R@50} 
& \textbf{Unique} \\

\midrule

ReDial~\cite{redial}         & \xmark & \xmark & \xmark & \cmark & \xmark & \textcolor{red}{3} & 2.4 & 14.0 & 32.0 & \_ & \_ & \_ & \_ & \_ \\ 

\hdashline 

KBRD~\cite{kbrd}             & \cmark & \xmark & \xmark & \cmark & \xmark & \textcolor{orange}{2} & 3.0 & 16.3 & 33.8 & \_ & \_ & \_ & \_ & \_ \\ 
KGSF~\cite{kgsf}             & \cmark & \xmark & \xmark & \cmark & \xmark & \textcolor{red}{3} & 3.9 & 18.3 & 37.8 & \_ & \_ & \_ & \_ & \_ \\ 
KECRS~\cite{kecrs}           & \cmark & \xmark & \xmark & \cmark & \xmark & \textcolor{orange}{2} & 2.3 & 15.7 & 36.6 & \_ & \_ & \_ & \_ & \_ \\ 
BARCOR~\cite{barcor}         & \cmark & \xmark & \xmark & \cmark & \xmark & \textcolor{green}{1} & 2.5 & 16.2 & 35.0 & \_ & \_ & \_ & \_ & \_ \\ 
UniCRS~\cite{unicrs}         & \cmark & \xmark & \xmark & \cmark & \cmark & \textcolor{red}{3} & 5.1 & 22.4 & 42.8 & \_ & \textbf{9.4} & \textbf{25.0} & \textbf{41.0} & \_ \\
RecInDial~\cite{recindial}   & \cmark & \xmark & \xmark & \cmark & \xmark & \textcolor{green}{1} & 3.1 & 14.0 & 27.0 & \_ & \_ & \_ & \_ & \_ \\ 
VRICR~\cite{vricr}           & \cmark & \xmark & \xmark & \cmark & \xmark & \textcolor{red}{3} & 5.7 & \underline{25.1} & 41.6 & \_ & \_ & \_ & \_ & \_ \\ 

\hdashline

RevCore~\cite{revcore}       & \cmark & \cmark & \xmark & \cmark & \xmark & \textcolor{orange}{2} & \textbf{6.1} & 23.6 & \underline{45.4} & \_ & \_ & \_ & \_ & \_ \\ 
C$^{2}$-CRS~\cite{c2crs}     & \cmark & \cmark & \xmark & \cmark & \xmark & \textcolor{orange}{2} & 5.3 & 23.3 & 40.7 & \_ & \_ & \_ & \_ & \_ \\ 

\hdashline

MESE~\cite{mese}             & \xmark & \xmark & \cmark & \cmark & \xmark &\textcolor{green}{1} & 5.6 & \textbf{25.6} & \textbf{45.5} & \_ & 4.8 & 13.5 & 30.1 & \_ \\
\textbf{PECRS-small}         & \xmark & \xmark & \cmark & \xmark & \cmark & \textcolor{green}{1} & 4.7 & 20.8 & 40.5 & \underline{463} & 5.4 & 16.1 & 33.3 & \underline{34} \\
\textbf{PECRS-medium}        & \xmark & \xmark & \cmark & \xmark & \cmark & \textcolor{green}{1} & \underline{5.8} & 22.5 & 41.6 & \textbf{634} & \underline{5.7} & \underline{17.9} & \underline{33.7} & \textbf{72} \\


\bottomrule

\end{tabular}
}
\caption{\small Results of the recommendation task compared with the state-of-the-art on ReDial and INSPIRED. Results are taken from respective papers. Best numbers are in \textbf{bold}, second best \underline{underlined}.}
\label{rec}
\end{table*}

\paragraph{Evaluation Metrics.} 
We follow the common practices~\cite{mese,unicrs} to evaluate PECRS on both recommendation performance and response generation quality. For recommendation subtask, we measure recall with \emph{Recall@K (R@K)} metric, taking $K\in\{1,10,50\}$. 
In order to assess the recommendation coverage, we also report the number of different items predicted by the model over the test set, denoted as \emph{Unique}. ReDial and INSPIRED contain 6,637 and 1,546 unique items in total (\Cref{datasets}) and 1,872 and 264 items in the test set, respectively.

We use both \emph{reference-based} and \emph{reference-free} metrics to evaluate response generation quality. For reference-based metrics, 
we adopt \emph{ROUGE@K (RG-K)}~\cite{rouge} with $K\in\{1,2\}$.
To verify whether the model could correctly predict a movie in response when required, we inspect the presence of the ``[ITEM]'' token in generated responses \emph{w.r.t.} ground truth requirement of movie prediction via \emph{F-1} score. For reference-free metrics, we use \emph{Perplexity (PPL)} to assess the text fluency and \emph{Distinct@K (Dist@K)} with $K\in\{2,3,4\}$ to measure the diversity of generated responses.

\paragraph{Implementation.} 
We choose GPT-2~\cite{gpt2} as the backbone LM, and experiment with two different model sizes, \ie GPT-2 small and GPT-2 medium, which enable us to compare against popular CRS approaches. Accordingly, we have \textbf{PECRS-small} and \textbf{PECRS-medium}.
We highlight that PECRS is flexible and can support other choices of decoder-only LMs. We use the public pre-trained checkpoints from HuggingFace \emph{transformers} library~\cite{transformers}. We set $M_{\text{train}}=150$ for training and $M_{\text{infer}}=700$ for inference. For ReDial, we train for $10$ epochs with effective batch size $8$; while for INSPIRED, we train for $20$ epochs with an effective batch size of $2$. Parameter optimization is performed by AdamW~\cite{adamw} with linear learning rate warmup strategy. We set maximum learning rate as $3e-5$ for PECRS-small and PECRS-medium and warmup for 1 epoch. During training, we balance losses with $\alpha=0.15$, $\beta=0.85$, and $\gamma=1.0$. We cap dialogue context length at $256$ tokens and response length at $64$ tokens.
We use checkpoint with the highest mean of R@1, R@10 and R@50 for inference. 
PECRS generates the response with top-k sampling, using $k=50$.
The movie item metadata is obtained from The Movie Database through \emph{tmdbv3api} library\footnote{https://github.com/AnthonyBloomer/tmdbv3api}. 

\begin{table}[]
\setlength{\tabcolsep}{5pt}
\resizebox{\columnwidth}{!}{  
\fontsize{20pt}{20pt}\selectfont
\begin{tabular}{lccccccc}

\toprule 

\multirow{2}{*}{\textbf{Model}} 
& \multicolumn{3}{c}{\textbf{Reference-based}}      
& \multicolumn{4}{c}{\textbf{Reference-free}} \\

\cmidrule(lr){2-4}
\cmidrule(lr){5-8}

& \textbf{RG-1} 
& \textbf{RG-2} 
& \textbf{F-1} 
& \textbf{PPL} 
& \textbf{Dist@2} 
& \textbf{Dist@3} 
& \textbf{Dist@4} \\

\midrule

C$^{2}$-CRS             & \_ & \_ & \_ & \_ & 0.163 & 0.291 & 0.417 \\
UniCRS                  & \_ & \_ & \_ & \_ & 0.492 & 0.648 & 0.832 \\
RecInDial               & \_ & \_ & \_ & \_ & 0.518 & 0.624 & 0.598 \\
MESE                    & \_ & \_ & \_ & 12.9 & \textbf{0.822} & 1.152 & 1.313 \\

\hdashline 

\textbf{PECRS-small}   & \underline{36.28} & \underline{14.77} & \underline{86.04} & \underline{9.89} & 0.745 & \underline{1.462} & \underline{2.132} \\
\textbf{PECRS-medium}  & \textbf{36.86} & \textbf{15.27} & \textbf{86.36} & \textbf{8.98} & \underline{0.820} & \textbf{1.552} & \textbf{2.154} \\

\bottomrule

\end{tabular}
}
\caption{\small Results of conversation task compared with the state-of-the-art on ReDial.}
\label{gen}
\end{table}

\begin{table}[]
\resizebox{\columnwidth}{!}{  
\begin{tabular}{lccc}
\toprule 

\textbf{Aspect} & \textbf{MESE} & \textbf{PECRS-small} & \textbf{Tie} \\

\midrule 

Fluency         & 28.00 (1.63) & \textbf{46.67} (5.91) & 25.33 (6.24) \\
Relevancy       & 26.33 (2.62) & \textbf{46.00} (0.82) & 27.67 (2.87) \\ 

\bottomrule 

\end{tabular}
}
\caption{\small Human evaluation on 100 random ReDial test data points. We show the average scores for three human raters, with standard deviation in parenthesis.}
\label{human}
\end{table}

\subsection{Comparison with State-of-the-Art }
\label{subsec:4.2}

The results on recommendation task are summarized in \Cref{rec}. Note that RevCore~\cite{revcore} and C$^2$CRS~\cite{c2crs} are not directly comparable to our method as they use additional movie review information. PECRS generally outperforms the baselines using KG and extra model, such as KGSF~\cite{kgsf} and UniCRS~\cite{unicrs}, on both datasets. Compared to the baselines with single training stage, PECRS surpasses BARCOR~\cite{barcor} and RecInDial~\cite{recindial}. MESE~\cite{mese} also uses the item descriptions and employs two additional modules to encode items. In contrast, our PECRS is simpler and more straightforward, and it is the first approach without using either KG or supplementary module, but only relying on the pre-trained LM. PECRS-medium outperforms MESE for Recall@1 on ReDial, achieving SOTA, and largely surpasses MESE for all metrics on INSPIRED.
Besides, PECRS-medium is superior to -small on all metrics, which demonstrates that fine-tuning a larger LM brings more gains thanks to its stronger representation ability.

\Cref{gen} summarizes the results on conversation task, where PECRS achieves promising performance on both types of metrics. Both PECRS-small and -medium surpass all baselines over Dist@3 and Dist@4. Comparing PECRS-small and -medium shows that Dist@K improvements can be achieved by scaling up the backbone model. Thus, we believe that larger LMs can bring better results, and fine-tuning them with plugin style to acquire CRS capability is a promising research direction. A human evaluation (\Cref{human}) for fluency and relevancy on ReDial test set with three volunteer graduate students with professional English proficiency confirms a preference for PECRS-small generated text over MESE outputs.

\subsection{Ablation Study}
\label{subsec:4.3}

\begin{table}[]
\setlength{\tabcolsep}{5pt}
\resizebox{\columnwidth}{!}{  
\fontsize{20pt}{20pt}\selectfont
\begin{tabular}{lccccc}

\toprule

\multirow{2}{*}{\textbf{Model}} 
& \multirow{2}{*}{\begin{tabular}[c]{@{}c@{}}\textbf{Time/}\\ \textbf{batch (s)} \end{tabular}}
& \multicolumn{2}{c}{\textbf{Rec.}}      
& \multicolumn{2}{c}{\textbf{Conv.}} \\

\cmidrule(lr){3-4}
\cmidrule(lr){5-6}

& & \textbf{R@50} 
& \textbf{Unique} 
& \textbf{RG-1} 
& \textbf{Dist@2} \\

\midrule

\textbf{PECRS-small}        & \underline{6.1} & \underline{40.5} & \underline{463} & 36.28 & 0.745 \\

\hdashline 

w/o Recall loss             & \underline{6.1} & 19.3 & 21 & \textbf{37.67} & 0.678 \\
w/o Rerank loss             & \underline{6.1} & 12.2 & 87 & 36.50 & 0.745 \\
w/o Generation loss         & \underline{6.1} & 39.2 & 451 & 7.76 & \textbf{11.907} \\
w/o Neg. sharing (batch)    & 8.6 & 39.8 & 291 & 36.40 & 0.747 \\
w/o Neg. sharing (tasks)    & 9.1 & \textbf{40.8} & 434 & 35.98 & 0.727 \\
w/o Item pooling            & \underline{6.1} & 39.6 & \textbf{530} & \underline{36.60} & 0.748 \\
w/o Item head               & \underline{6.1} & 37.9 & 453 & 36.33 & 0.726 \\
w/o Metadata (just title)   & \textbf{4.2} & 35.8 & 384 & 36.38 & \underline{0.765} \\

\bottomrule 

\end{tabular}
}
\caption{\small Models comparison with different modules and optimization strategies on ReDial with PECRS-small.}
\label{ablation}
\end{table}








\begin{table}[]
\setlength{\tabcolsep}{5pt}
\resizebox{\columnwidth}{!}{  
\fontsize{18pt}{18pt}\selectfont
\begin{tabular}{lcccccc}

\toprule 

\textbf{Removed} 
& None 
& Title 
& Actor(s) 
& Director(s) 
& Genre(s) 
& Plot \\

\midrule 

\textbf{R@50}     & \textbf{33.3} & 29.8 & 26.9 & \underline{32.5} & 30.5 & 20.7 \\

\bottomrule 

\end{tabular}
}
\caption{\small Effect of pruning fields of items metadata at inference on INSPIRED with PECRS-small.}
\label{metadata}
\end{table}

We also conduct ablative experiments to analyze the architecture and optimization design of PECRS.
Reported in \Cref{ablation}, all the components and training strategies contribute to the performance gains on both recommendation and conversation tasks. In particular, recommendation collapses without either loss from its two-stage processes, \ie retrieval and re-ranking ; and suffers without the generation loss. Sharing negative samples across batch elements and tasks leads to significant improvements on training efficiency and marginal gains on recommendation performance.

In \Cref{metadata}, we conduct a further ablation on the textual fields within items description. We observe that every field contributes to the recommendation performance, especially the plot. This suggests that richer metadata would yield even more recall gains.

\subsection{Comparison with Large Language Models}
\label{subsec:4.4}

\begin{table}[]
\setlength{\tabcolsep}{5pt}
\resizebox{\columnwidth}{!}{  
\fontsize{20pt}{20pt}\selectfont
\begin{tabular}{lcccccc}

\toprule

\multirow{2}{*}{\textbf{Model}} 
& \multicolumn{4}{c}{\textbf{Rec.}}      
& \multicolumn{2}{c}{\textbf{Conv.}} \\

\cmidrule(lr){2-5}
\cmidrule(lr){6-7}

& \textbf{R@1} 
& \textbf{R@10} 
& \textbf{R@50} 
& \textbf{Unique} 
& \textbf{RG-1} 
& \textbf{RG-2} \\

\midrule

PECRS-small                 & 5.4 & \textbf{16.1} & \textbf{33.3} & \textbf{34} & \textbf{29.72} & \textbf{8.26} \\

\hdashline 

Llama-2-7B-chat             & \textbf{9.3} & \emph{9.3} & \emph{9.3} & 26 & 19.88 & 2.88 \\
Vicuna-1.5-7B               & 8.2 & \emph{8.2} & \emph{8.2} & 23 & 21.18 & 3.50 \\

\bottomrule 

\end{tabular}
}
\caption{\small Comparison between PECRS-small and two popular LLMs in zero-shot on INSPIRED test set.}
\label{llms}
\end{table}

\begin{table}[]
\setlength{\tabcolsep}{5pt}
\resizebox{\columnwidth}{!}{  
\fontsize{20pt}{20pt}\selectfont
\begin{tabular}{lccccc}

\toprule 

\multirow{2}{*}{\textbf{Decoding Strategy}} 
& \multicolumn{2}{c}{\textbf{Reference-based}} 
& \multicolumn{3}{c}{\textbf{Reference-free}} \\

\cmidrule(lr){2-3}
\cmidrule(lr){4-6}

& \textbf{RG-1}         
& \textbf{RG-2}        
& \textbf{Dist@2} 
& \textbf{Dist@3} 
& \textbf{Dist@4} \\

\midrule

Greedy decoding                             & 38.54 & 16.25 & 0.208 & 0.311 & 0.390 \\

\hdashline 

Beam search                                 & 38.23 & 16.83 & 0.235 & 0.353 & 0.444 \\

\hdashline 

Diverse beam search (diversity=0.5)         & 39.94 & \underline{17.30} & 0.190 & 0.287 & 0.361 \\
Diverse beam search (diversity=1.0)         & \textbf{40.29} & \textbf{17.40} & 0.179 & 0.264 & 0.320 \\
Diverse beam search (diversity=1.5)         & \underline{40.07} & 17.23 & 0.172 & 0.246 & 0.290 \\

\hdashline 

Top-k sampling (k=25)                       & 33.54 & 14.40 & 0.593 & 1.177 & 1.806 \\
Top-k sampling (k=50)                       & 33.37 & 14.17 & \textbf{0.647} & \underline{1.300} & \underline{1.989} \\  
Top-k sampling (k=75)                       & 33.48 & 14.15 & \underline{0.644} & \textbf{1.303} & \textbf{1.992} \\  

\hdashline 

Nucleus sampling (p=0.90)                   & 36.35 & 16.04 & 0.329 & 0.555 & 0.760 \\
Nucleus sampling (p=0.95)                   & 36.44 & 16.02 & 0.351 & 0.594 & 0.804 \\
Nucleus sampling (p=0.99)                   & 36.60 & 16.07 & 0.352 & 0.593 & 0.809 \\

\bottomrule 

\end{tabular}
}
\caption{\small The conversation performance of PECRS-small with different decoding strategies on ReDial. Except \emph{Greedy decoding}, all other techniques use a beam width of 4.}
\label{decoding}
\end{table}

Lastly, we compare our fine-tuning approach with Large Language Models (LLMs). Instruction-tuned LLMs have brought a seismic shift in NLP recently, due to their ability to seamlessly conduct many tasks in a zero-shot fashion through prompts, by-passing the need for task-specific supervised fine-tuning \citep{t0,flan,instructgpt}, including in recommender systems \citep{hou2023large}.

We use two popular LLMs: Llama-2-7B-chat\footnote{https://huggingface.co/meta-llama/Llama-2-7b-chat-hf} \citep{llama2}, and Vicuna-1.5-7B\footnote{https://huggingface.co/lmsys/vicuna-7b-v1.5} \citep{vicuna}. For each model, we condition on the context, and prompt the LLM to predict the Recommender response, which should include a movie name. We infer in \emph{bfloat16}, decode with greedy decoding, and check if the ground-truth movie name is included in the generated response. As seen in \Cref{llms}, the conversational recommendation capability of LLMs in zero-shot is very promising, as they outperform PECRS-small in Recall@1 on INSPIRED. However, due to the lack of a dedicated recommendation module, LLMs used in this fashion cannot suggest a full list of items, hence their recall plateaus at the Recall@1 value. They also tend to recommend fewer different movies (lower \textbf{Unique}). Exploring the ranking of a larger list of recommended items with LLMs is a promising future research avenue.

\section{Analysis}
\label{sec:5}
In this section, we provide more detailed insights about the behavior of PECRS.

\subsection{Conversation Evaluation}
\label{subsec:5.1}

\begin{figure}
    \includegraphics[width=\columnwidth]{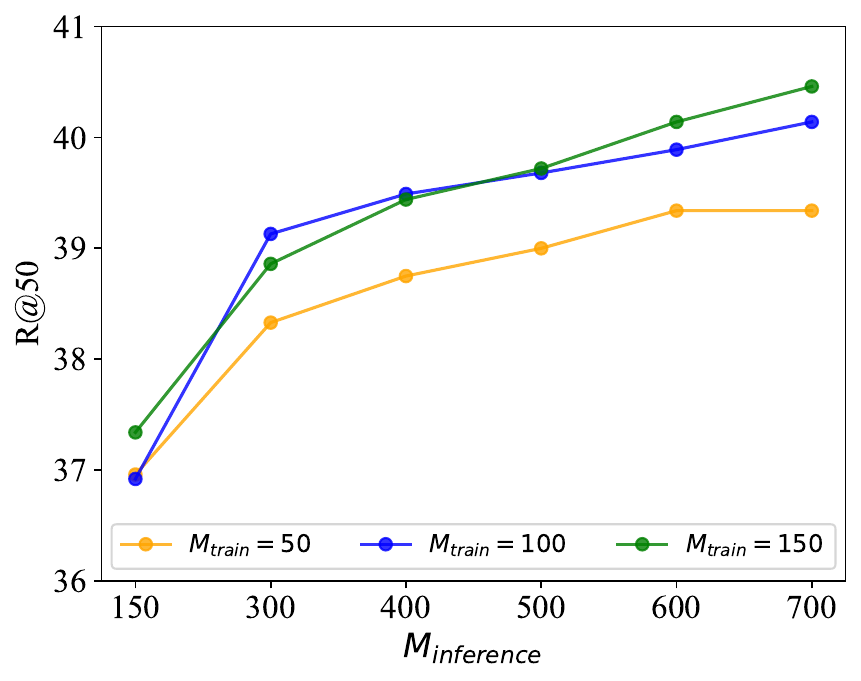}
    \caption{\small The R@50 results of PECRS-small using the different $M_{\text{train}}$ and $M_{\text{inference}}$ pairs on ReDial dataset.}
    \label{fig:3}
\end{figure}

We first study the effects of different LM's decoding strategies on conversational performance over Dist@K metric. Specifically, we analyze the greedy decoding, beam search, diverse beam search~\cite{dbs}, top-k sampling~\cite{topk} and nucleus sampling~\cite{topp} strategies on PECRS-small. Reported in \Cref{decoding}, reference-based metrics (RG-K) show much less variance on different decoding strategies compared to the reference-free metrics (Dist@K). Meanwhile, the correlation between reference-based and reference-free metrics is weak under different decoding strategies. Moreover, PECRS without training for generation can achieve $11.907$ on Dist@2 metric (see \emph{w/o Generation loss} in \Cref{ablation}), but merely $7.76$ on RG-1 metric. \textit{This observation implies that Dist@K metrics are not reliable to evaluate the quality of response generation}. Since Dist@K metrics have become the most popular choice in evaluating conversation performance of CRS~\cite{c2crs,unicrs,mese}, we advocate for applying other metrics, in particular reference-based metrics including n-gram overlap like ROUGE or semantic similarity like BERTScore \cite{zhang2019bertscore}, to provide more accurate evaluation on the response generation of CRS.

\subsection{Negative Sampling}
\label{subsec:5.2}

Now we analyze how the hyper-parameters of negative sampling, \ie $M_{\text{train}}$ and $M_{\text{inference}}$, affect the recommendation performance. \Cref{fig:3} illustrates the results of different choices of $M_{\text{train}}$ and $M_{\text{inference}}$ pairs. In general, $M_{\text{train}}$ and $M_{\text{inference}}$ have significant impacts on the recommendation performance, and larger $M_{\text{train}}$ and $M_{\text{inference}}$ lead to better results. However, increasing $M$ will reduce the training and inference efficiency. Thus, there is a trade-off between efficiency and recommendation performance for the selection of $M$.

\begin{figure}
    \includegraphics[width=\columnwidth]{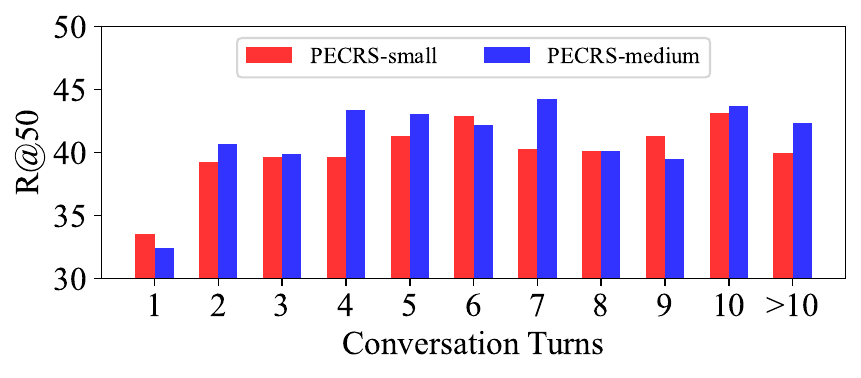}
    \caption{\small R@50 of PECRS on ReDial per number of conversation turns prior to the CRS response.}
    \label{fig:4}
\end{figure}

\subsection{Conversation Turns}
\label{subsec:5.3}

Lastly, we investigate how robust is PECRS with regards to the richness of dialogue context. In \Cref{fig:4}, we group data points by number of utterances happening before the CRS response.
We observe that PECRS performs well in recommendation for a wide range of context length, with only a moderate drop when there is only one prior utterance.

\section{Conclusion}
\label{sec:6}
In this work, we formulate conversational recommendation as a language processing task and propose a unified parameter-efficient CRS (PECRS) framework to solve it in a single-stage end-to-end manner. PECRS effectively addresses the inferior training efficiency via parameter-efficient fine-tuning techniques and semantic misalignment issues via joint conversation and recommendation modeling. Through experiments, we show that PECRS achieves performance competitive with SOTA on both recommendation and response generation on benchmark datasets. 
Moreover, for response evaluation, we reveal the commonly used Dist@K metrics are not reliable, and advocate for reference-based metrics (e.g ROUGE) for more accurate evaluation. Generally, we show that it is promising to explore unified framework for CRS under the natural language paradigm via language model and rich textual items data. 

\section*{Limitations}

Our work adheres to standard practices for dataset construction and model evaluation. However, we acknowledge three limitations: (1) Recommender utterances containing multiple items are separated into individual data points, which is sub-optimal as the model may only be accurate for the top-ranked item in each data point. (2) If we train PECRS to predict multiple items within the same utterance, it is challenging to compare with current methods, as they do not make simultaneous predictions. (3) All items mentioned by the recommender are considered recommendations, although some may be references to previous discussions or express dislikes rather than actual recommendations. 

The maximum context length for the backbone LM is another limitation. We have demonstrated that increasing $M_{\text{inference}}$ yields better recommendation performance (ref. \Cref{subsec:5.2}). However, we are constrained by the maximum input length of 1024 for GPT-2, which limits the candidate set size after concatenating with dialogue context. The potential extensions may involve performing inference with multiple forward passes to score batches of $M_{\text{inference}}$ items, or using a backbone that supports longer input lengths, albeit at a higher computational cost.
We only experiment with relatively small backbone, \ie GPT2-small and -medium, due to resource limitation. However, PECRS is flexible and can be seamlessly applied to larger backbones like LLaMA~\cite{llama}.


\bibliography{anthology,custom}

\begin{thebibliography}{64}
\expandafter\ifx\csname natexlab\endcsname\relax\def\natexlab#1{#1}\fi

\bibitem[{Auer et~al.(2007)Auer, Bizer, Kobilarov, Lehmann, Cyganiak, and Ives}]{dbpedia}
S\"{o}ren Auer, Christian Bizer, Georgi Kobilarov, Jens Lehmann, Richard Cyganiak, and Zachary Ives. 2007.
\newblock \href {https://link.springer.com/chapter/10.1007/978-3-540-76298-0_52} {Dbpedia: A nucleus for a web of open data}.
\newblock page 722–735. Springer-Verlag.

\bibitem[{Chen et~al.(2019)Chen, Lin, Zhang, Ding, Cen, Yang, and Tang}]{kbrd}
Qibin Chen, Junyang Lin, Yichang Zhang, Ming Ding, Yukuo Cen, Hongxia Yang, and Jie Tang. 2019.
\newblock \href {https://doi.org/10.18653/v1/D19-1189} {Towards knowledge-based recommender dialog system}.
\newblock In \emph{Proceedings of the 2019 Conference on Empirical Methods in Natural Language Processing and the 9th International Joint Conference on Natural Language Processing (EMNLP-IJCNLP)}, pages 1803--1813, Hong Kong, China. Association for Computational Linguistics.

\bibitem[{Chen et~al.(2023)Chen, Duan, Wang, He, Lu, Dai, and Qiao}]{chen2023vision}
Zhe Chen, Yuchen Duan, Wenhai Wang, Junjun He, Tong Lu, Jifeng Dai, and Yu~Qiao. 2023.
\newblock \href {https://openreview.net/forum?id=plKu2GByCNW} {Vision transformer adapter for dense predictions}.
\newblock In \emph{The Eleventh International Conference on Learning Representations}.

\bibitem[{Chiang et~al.(2023)Chiang, Li, Lin, Sheng, Wu, Zhang, Zheng, Zhuang, Zhuang, Gonzalez et~al.}]{vicuna}
Wei-Lin Chiang, Zhuohan Li, Zi~Lin, Ying Sheng, Zhanghao Wu, Hao Zhang, Lianmin Zheng, Siyuan Zhuang, Yonghao Zhuang, Joseph~E Gonzalez, et~al. 2023.
\newblock Vicuna: An open-source chatbot impressing gpt-4 with 90\%* chatgpt quality.
\newblock \emph{See https://vicuna. lmsys. org (accessed 14 April 2023)}.

\bibitem[{Christakopoulou et~al.(2016)Christakopoulou, Radlinski, and Hofmann}]{crs}
Konstantina Christakopoulou, Filip Radlinski, and Katja Hofmann. 2016.
\newblock \href {https://doi.org/10.1145/2939672.2939746} {Towards conversational recommender systems}.
\newblock In \emph{Proceedings of the 22nd ACM SIGKDD International Conference on Knowledge Discovery and Data Mining}, page 815–824. Association for Computing Machinery.

\bibitem[{Deng et~al.(2023)Deng, Zhang, Xu, Lei, Chua, and Lam}]{unimind}
Yang Deng, Wenxuan Zhang, Weiwen Xu, Wenqiang Lei, Tat-Seng Chua, and Wai Lam. 2023.
\newblock \href {https://doi.org/10.1145/3570640} {A unified multi-task learning framework for multi-goal conversational recommender systems}.
\newblock \emph{ACM Trans. Inf. Syst.}, 41(3).

\bibitem[{Dettmers et~al.(2023)Dettmers, Pagnoni, Holtzman, and Zettlemoyer}]{qlora}
Tim Dettmers, Artidoro Pagnoni, Ari Holtzman, and Luke Zettlemoyer. 2023.
\newblock \href {https://arxiv.org/pdf/2305.14314.pdf} {Qlora: Efficient finetuning of quantized llms}.
\newblock \emph{ArXiv}, abs/2305.14314.

\bibitem[{Fan et~al.(2018)Fan, Lewis, and Dauphin}]{topk}
Angela Fan, Mike Lewis, and Yann Dauphin. 2018.
\newblock \href {https://doi.org/10.18653/v1/P18-1082} {Hierarchical neural story generation}.
\newblock In \emph{Proceedings of the 56th Annual Meeting of the Association for Computational Linguistics (Volume 1: Long Papers)}, pages 889--898. Association for Computational Linguistics.

\bibitem[{Fu et~al.(2023)Fu, Yuan, Song, Yuan, Cheng, Cheng, Zhang, Wang, and Pan}]{Fu2023ExploringAT}
Junchen Fu, Fajie Yuan, Yu~Song, Zheng Yuan, Mingyue Cheng, Shenghui Cheng, Jiaqi Zhang, Jie Wang, and Yunzhu Pan. 2023.
\newblock \href {https://arxiv.org/pdf/2305.15036.pdf} {Exploring adapter-based transfer learning for recommender systems: Empirical studies and practical insights}.
\newblock \emph{ArXiv}, abs/2305.15036.

\bibitem[{Gao et~al.(2021)Gao, Lei, He, de~Rijke, and Chua}]{survey-gao}
Chongming Gao, Wenqiang Lei, Xiangnan He, M.~de~Rijke, and Tat-Seng Chua. 2021.
\newblock \href {https://arxiv.org/pdf/2101.09459.pdf} {Advances and challenges in conversational recommender systems: A survey}.
\newblock \emph{AI Open}, 2:100--126.

\bibitem[{Gutmann and Hyv\"{a}rinen(2012)}]{nce}
Michael~U. Gutmann and Aapo Hyv\"{a}rinen. 2012.
\newblock \href {https://jmlr.org/papers/volume13/gutmann12a/gutmann12a.pdf} {Noise-contrastive estimation of unnormalized statistical models, with applications to natural image statistics}.
\newblock \emph{J. Mach. Learn. Res.}, 13:307–361.

\bibitem[{Hayati et~al.(2020)Hayati, Kang, Zhu, Shi, and Yu}]{inspired}
Shirley~Anugrah Hayati, Dongyeop Kang, Qingxiaoyang Zhu, Weiyan Shi, and Zhou Yu. 2020.
\newblock \href {https://doi.org/10.18653/v1/2020.emnlp-main.654} {{INSPIRED}: Toward sociable recommendation dialog systems}.
\newblock In \emph{Proceedings of the 2020 Conference on Empirical Methods in Natural Language Processing (EMNLP)}, pages 8142--8152, Online. Association for Computational Linguistics.

\bibitem[{He et~al.(2022)He, Li, Zhang, Yang, and Wang}]{he2022parameterefficient}
Xuehai He, Chunyuan Li, Pengchuan Zhang, Jianwei Yang, and Xin~Eric Wang. 2022.
\newblock \href {https://arxiv.org/pdf/2203.16329.pdf} {Parameter-efficient model adaptation for vision transformers}.
\newblock \emph{ArXiv}, abs/2203.16329.

\bibitem[{Holtzman et~al.(2020)Holtzman, Buys, Du, Forbes, and Choi}]{topp}
Ari Holtzman, Jan Buys, Li~Du, Maxwell Forbes, and Yejin Choi. 2020.
\newblock \href {https://openreview.net/forum?id=rygGQyrFvH} {The curious case of neural text degeneration}.
\newblock In \emph{International Conference on Learning Representations}.

\bibitem[{Hou et~al.(2023)Hou, Zhang, Lin, Lu, Xie, McAuley, and Zhao}]{hou2023large}
Yupeng Hou, Junjie Zhang, Zihan Lin, Hongyu Lu, Ruobing Xie, Julian McAuley, and Wayne~Xin Zhao. 2023.
\newblock Large language models are zero-shot rankers for recommender systems.
\newblock \emph{arXiv preprint arXiv:2305.08845}.

\bibitem[{Houlsby et~al.(2019)Houlsby, Giurgiu, Jastrzebski, Morrone, De~Laroussilhe, Gesmundo, Attariyan, and Gelly}]{adapters}
Neil Houlsby, Andrei Giurgiu, Stanislaw Jastrzebski, Bruna Morrone, Quentin De~Laroussilhe, Andrea Gesmundo, Mona Attariyan, and Sylvain Gelly. 2019.
\newblock \href {https://proceedings.mlr.press/v97/houlsby19a.html} {Parameter-efficient transfer learning for {NLP}}.
\newblock In \emph{Proceedings of the 36th International Conference on Machine Learning}, volume~97, pages 2790--2799.

\bibitem[{Hu et~al.(2022{\natexlab{a}})Hu, Huang, Zhang, and Liu}]{crif}
Chenhao Hu, Shuhua Huang, Yansen Zhang, and Yubao Liu. 2022{\natexlab{a}}.
\newblock \href {https://doi.org/10.1145/3477495.3531844} {Learning to infer user implicit preference in conversational recommendation}.
\newblock In \emph{Proceedings of the 45th International ACM SIGIR Conference on Research and Development in Information Retrieval}, page 256–266. Association for Computing Machinery.

\bibitem[{Hu et~al.(2022{\natexlab{b}})Hu, yelong shen, Wallis, Allen-Zhu, Li, Wang, Wang, and Chen}]{lora}
Edward~J Hu, yelong shen, Phillip Wallis, Zeyuan Allen-Zhu, Yuanzhi Li, Shean Wang, Lu~Wang, and Weizhu Chen. 2022{\natexlab{b}}.
\newblock \href {https://openreview.net/forum?id=nZeVKeeFYf9} {Lo{RA}: Low-rank adaptation of large language models}.
\newblock In \emph{International Conference on Learning Representations}.

\bibitem[{Hu et~al.(2023)Hu, Lan, Wang, Xu, Lim, Lee, Bing, Xu, and Poria}]{llmadapter}
Zhiqiang Hu, Yihuai Lan, Lei Wang, Wanyu Xu, Ee-Peng Lim, Roy Ka-Wei Lee, Lidong Bing, Xing Xu, and Soujanya Poria. 2023.
\newblock \href {https://arxiv.org/pdf/2304.01933.pdf} {Llm-adapters: An adapter family for parameter-efficient fine-tuning of large language models}.
\newblock \emph{ArXiv}, abs/2304.01933.

\bibitem[{Jannach et~al.(2021)Jannach, Manzoor, Cai, and Chen}]{survey-jannach}
Dietmar Jannach, Ahtsham Manzoor, Wanling Cai, and Li~Chen. 2021.
\newblock \href {https://doi.org/10.1145/3453154} {A survey on conversational recommender systems}.
\newblock \emph{ACM Comput. Surv.}, 54(5).

\bibitem[{Lei et~al.(2020)Lei, He, Miao, Wu, Hong, Kan, and Chua}]{ear}
Wenqiang Lei, Xiangnan He, Yisong Miao, Qingyun Wu, Richang Hong, {Min Yen} Kan, and {Tat Seng} Chua. 2020.
\newblock \href {https://doi.org/10.1145/3336191.3371769} {Estimation–action–reflection: Towards deep interaction between conversational and recommender systems}.
\newblock In \emph{WSDM 2020 - Proceedings of the 13th International Conference on Web Search and Data Mining}, pages 304--312. Association for Computing Machinery, Inc.

\bibitem[{Lester et~al.(2021)Lester, Al-Rfou, and Constant}]{softprompt}
Brian Lester, Rami Al-Rfou, and Noah Constant. 2021.
\newblock \href {https://doi.org/10.18653/v1/2021.emnlp-main.243} {The power of scale for parameter-efficient prompt tuning}.
\newblock In \emph{Proceedings of the 2021 Conference on Empirical Methods in Natural Language Processing}, pages 3045--3059, Online and Punta Cana, Dominican Republic. Association for Computational Linguistics.

\bibitem[{Lewis et~al.(2020)Lewis, Liu, Goyal, Ghazvininejad, Mohamed, Levy, Stoyanov, and Zettlemoyer}]{bart}
Mike Lewis, Yinhan Liu, Naman Goyal, Marjan Ghazvininejad, Abdelrahman Mohamed, Omer Levy, Veselin Stoyanov, and Luke Zettlemoyer. 2020.
\newblock \href {https://doi.org/10.18653/v1/2020.acl-main.703} {{BART}: Denoising sequence-to-sequence pre-training for natural language generation, translation, and comprehension}.
\newblock In \emph{Proceedings of the 58th Annual Meeting of the Association for Computational Linguistics}, pages 7871--7880, Online. Association for Computational Linguistics.

\bibitem[{Li et~al.(2018)Li, Kahou, Schulz, Michalski, Charlin, and Pal}]{redial}
Raymond Li, Samira Kahou, Hannes Schulz, Vincent Michalski, Laurent Charlin, and Chris Pal. 2018.
\newblock \href {https://dl.acm.org/doi/pdf/10.5555/3327546.3327641} {Towards deep conversational recommendations}.
\newblock In \emph{Proceedings of the 32nd International Conference on Neural Information Processing Systems}, page 9748–9758. Curran Associates Inc.

\bibitem[{Li et~al.(2022)Li, Xie, Zhu, Ao, Zhuang, and He}]{uccr}
Shuokai Li, Ruobing Xie, Yongchun Zhu, Xiang Ao, Fuzhen Zhuang, and Qing He. 2022.
\newblock \href {https://doi.org/10.1145/3477495.3532074} {User-centric conversational recommendation with multi-aspect user modeling}.
\newblock In \emph{Proceedings of the 45th International ACM SIGIR Conference on Research and Development in Information Retrieval}, page 223–233. Association for Computing Machinery.

\bibitem[{Liang et~al.(2021)Liang, Hu, Xu, Miao, He, Chen, Geng, Liang, and Jiang}]{ntrd}
Zujie Liang, Huang Hu, Can Xu, Jian Miao, Yingying He, Yining Chen, Xiubo Geng, Fan Liang, and Daxin Jiang. 2021.
\newblock Learning neural templates for recommender dialogue system.
\newblock \emph{arXiv preprint arXiv:2109.12302}.

\bibitem[{Lin(2004)}]{rouge}
Chin-Yew Lin. 2004.
\newblock \href {https://aclanthology.org/W04-1013} {{ROUGE}: A package for automatic evaluation of summaries}.
\newblock In \emph{Text Summarization Branches Out}, pages 74--81, Barcelona, Spain. Association for Computational Linguistics.

\bibitem[{Lin et~al.(2023)Lin, Dai, Xi, Liu, Chen, Li, Zhu, Guo, Yu, Tang, and Zhang}]{lin2023howcr}
Jianghao Lin, Xinyi Dai, Yunjia Xi, Weiwen Liu, Bo~Chen, Xiangyang Li, Chenxu Zhu, Huifeng Guo, Yong Yu, Ruiming Tang, and Weinan Zhang. 2023.
\newblock \href {https://arxiv.org/pdf/2306.05817.pdf} {How can recommender systems benefit from large language models: A survey}.
\newblock \emph{ArXiv}, abs/2306.05817.

\bibitem[{Liu et~al.(2020)Liu, Wang, Niu, Wu, Che, and Liu}]{durecdial}
Zeming Liu, Haifeng Wang, Zheng-Yu Niu, Hua Wu, Wanxiang Che, and Ting Liu. 2020.
\newblock \href {https://doi.org/10.18653/v1/2020.acl-main.98} {Towards conversational recommendation over multi-type dialogs}.
\newblock In \emph{Proceedings of the 58th Annual Meeting of the Association for Computational Linguistics}, pages 1036--1049, Online. Association for Computational Linguistics.

\bibitem[{Liu et~al.(2023)Liu, Zhou, Liu, Wang, Niu, Wu, Che, Liu, and Xiong}]{mgcg}
Zeming Liu, Ding Zhou, Hao Liu, Haifeng Wang, Zheng-Yu Niu, Hua Wu, Wanxiang Che, Ting Liu, and Hui Xiong. 2023.
\newblock \href {https://doi.org/10.1109/TKDE.2022.3147210} {Graph-grounded goal planning for conversational recommendation}.
\newblock \emph{IEEE Transactions on Knowledge and Data Engineering}, 35(5):4923--4939.

\bibitem[{Loshchilov and Hutter(2019)}]{adamw}
Ilya Loshchilov and Frank Hutter. 2019.
\newblock \href {https://arxiv.org/pdf/1711.05101.pdf} {Decoupled weight decay regularization}.
\newblock \emph{ArXiv}, abs/1711.05101.

\bibitem[{Lu et~al.(2021)Lu, Bao, Song, Ma, Cui, Wu, and He}]{revcore}
Yu~Lu, Junwei Bao, Yan Song, Zichen Ma, Shuguang Cui, Youzheng Wu, and Xiaodong He. 2021.
\newblock \href {https://doi.org/10.18653/v1/2021.findings-acl.99} {{R}ev{C}ore: Review-augmented conversational recommendation}.
\newblock In \emph{Findings of the Association for Computational Linguistics: ACL-IJCNLP 2021}, pages 1161--1173, Online. Association for Computational Linguistics.

\bibitem[{Ma et~al.(2020)Ma, Takanobu, and Huang}]{crwalker}
Wenchang Ma, Ryuichi Takanobu, and Minlie Huang. 2020.
\newblock Cr-walker: Tree-structured graph reasoning and dialog acts for conversational recommendation.
\newblock \emph{arXiv preprint arXiv:2010.10333}.

\bibitem[{Mnih and Kavukcuoglu(2013)}]{nce3}
Andriy Mnih and Koray Kavukcuoglu. 2013.
\newblock \href {https://proceedings.neurips.cc/paper_files/paper/2013/file/db2b4182156b2f1f817860ac9f409ad7-Paper.pdf} {Learning word embeddings efficiently with noise-contrastive estimation}.
\newblock In \emph{Advances in Neural Information Processing Systems}, volume~26. Curran Associates, Inc.

\bibitem[{Mnih and Teh(2012)}]{nce2}
Andriy Mnih and Yee~Whye Teh. 2012.
\newblock \href {https://arxiv.org/pdf/1206.6426.pdf} {A fast and simple algorithm for training neural probabilistic language models}.
\newblock In \emph{Proceedings of the 29th International Coference on International Conference on Machine Learning}, page 419–426.

\bibitem[{Ouyang et~al.(2022)Ouyang, Wu, Jiang, Almeida, Wainwright, Mishkin, Zhang, Agarwal, Slama, Ray et~al.}]{instructgpt}
Long Ouyang, Jeffrey Wu, Xu~Jiang, Diogo Almeida, Carroll Wainwright, Pamela Mishkin, Chong Zhang, Sandhini Agarwal, Katarina Slama, Alex Ray, et~al. 2022.
\newblock Training language models to follow instructions with human feedback.
\newblock \emph{Advances in Neural Information Processing Systems}, 35:27730--27744.

\bibitem[{Pramod and Bafna(2022)}]{survey-pramod}
Dhanya Pramod and Prafulla Bafna. 2022.
\newblock \href {https://doi.org/10.1016/j.eswa.2022.117539} {Conversational recommender systems techniques, tools, acceptance, and adoption: A state of the art review}.
\newblock \emph{Expert Syst. Appl.}, 203(C).

\bibitem[{Radford et~al.(2019)Radford, Wu, Child, Luan, Amodei, and Sutskever}]{gpt2}
Alec Radford, Jeff Wu, Rewon Child, David Luan, Dario Amodei, and Ilya Sutskever. 2019.
\newblock \href {https://d4mucfpksywv.cloudfront.net/better-language-models/language-models.pdf} {Language models are unsupervised multitask learners}.

\bibitem[{Ren et~al.(2020)Ren, Yin, Chen, Wang, Hung, Huang, and Zhang}]{crsal}
Xuhui Ren, Hongzhi Yin, Tong Chen, Hao Wang, Nguyen Quoc~Viet Hung, Zi~Huang, and Xiangliang Zhang. 2020.
\newblock \href {https://doi.org/10.1145/3394592} {Crsal: Conversational recommender systems with adversarial learning}.
\newblock \emph{ACM Trans. Inf. Syst.}, 38(4).

\bibitem[{Sanh et~al.(2019)Sanh, Debut, Chaumond, and Wolf}]{distilbert}
Victor Sanh, Lysandre Debut, Julien Chaumond, and Thomas Wolf. 2019.
\newblock \href {https://arxiv.org/pdf/1910.01108.pdf} {Distilbert, a distilled version of bert: smaller, faster, cheaper and lighter}.
\newblock \emph{ArXiv}, abs/1910.01108.

\bibitem[{Sanh et~al.(2021)Sanh, Webson, Raffel, Bach, Sutawika, Alyafeai, Chaffin, Stiegler, Scao, Raja et~al.}]{t0}
Victor Sanh, Albert Webson, Colin Raffel, Stephen~H Bach, Lintang Sutawika, Zaid Alyafeai, Antoine Chaffin, Arnaud Stiegler, Teven~Le Scao, Arun Raja, et~al. 2021.
\newblock Multitask prompted training enables zero-shot task generalization.
\newblock \emph{arXiv preprint arXiv:2110.08207}.

\bibitem[{Schlichtkrull et~al.(2018)Schlichtkrull, Kipf, Bloem, van~den Berg, Titov, and Welling}]{rgcn}
Michael Schlichtkrull, Thomas~N. Kipf, Peter Bloem, Rianne van~den Berg, Ivan Titov, and Max Welling. 2018.
\newblock \href {https://arxiv.org/pdf/1703.06103.pdf} {Modeling relational data with graph convolutional networks}.
\newblock In \emph{The Semantic Web}, pages 593--607. Springer International Publishing.

\bibitem[{Speer et~al.(2017)Speer, Chin, and Havasi}]{conceptnet}
Robyn Speer, Joshua Chin, and Catherine Havasi. 2017.
\newblock \href {https://doi.org/10.1609/aaai.v31i1.11164} {Conceptnet 5.5: An open multilingual graph of general knowledge}.
\newblock In \emph{Proceedings of the AAAI Conference on Artificial Intelligence}.

\bibitem[{Sun and Zhang(2018)}]{crs2}
Yueming Sun and Yi~Zhang. 2018.
\newblock \href {https://doi.org/10.1145/3209978.3210002} {Conversational recommender system}.
\newblock In \emph{The 41st International ACM SIGIR Conference on Research \& Development in Information Retrieval}, page 235–244. Association for Computing Machinery.

\bibitem[{Touvron et~al.(2023{\natexlab{a}})Touvron, Lavril, Izacard, Martinet, Lachaux, Lacroix, Rozi{\`e}re, Goyal, Hambro, Azhar, Rodriguez, Joulin, Grave, and Lample}]{llama}
Hugo Touvron, Thibaut Lavril, Gautier Izacard, Xavier Martinet, Marie-Anne Lachaux, Timoth{\'e}e Lacroix, Baptiste Rozi{\`e}re, Naman Goyal, Eric Hambro, Faisal Azhar, Aur'elien Rodriguez, Armand Joulin, Edouard Grave, and Guillaume Lample. 2023{\natexlab{a}}.
\newblock \href {https://arxiv.org/abs/2302.13971v1} {Llama: Open and efficient foundation language models}.
\newblock \emph{ArXiv}, abs/2302.13971.

\bibitem[{Touvron et~al.(2023{\natexlab{b}})Touvron, Martin, Stone, Albert, Almahairi, Babaei, Bashlykov, Batra, Bhargava, Bhosale et~al.}]{llama2}
Hugo Touvron, Louis Martin, Kevin Stone, Peter Albert, Amjad Almahairi, Yasmine Babaei, Nikolay Bashlykov, Soumya Batra, Prajjwal Bhargava, Shruti Bhosale, et~al. 2023{\natexlab{b}}.
\newblock Llama 2: Open foundation and fine-tuned chat models.
\newblock \emph{arXiv preprint arXiv:2307.09288}.

\bibitem[{Vijayakumar et~al.(2018)Vijayakumar, Cogswell, Selvaraju, Sun, Lee, Crandall, and Batra}]{dbs}
Ashwin~K Vijayakumar, Michael Cogswell, Ramprasath~R. Selvaraju, Qing Sun, Stefan Lee, David Crandall, and Dhruv Batra. 2018.
\newblock \href {https://arxiv.org/pdf/1610.02424.pdf} {Diverse beam search: Decoding diverse solutions from neural sequence models}.
\newblock In \emph{Proceedings of the AAAI Conference on Artificial Intelligence}.

\bibitem[{Wang et~al.(2022{\natexlab{a}})Wang, Hu, Sha, Xu, Jiang, and Wong}]{recindial}
Lingzhi Wang, Huang Hu, Lei Sha, Can Xu, Daxin Jiang, and Kam-Fai Wong. 2022{\natexlab{a}}.
\newblock \href {https://aclanthology.org/2022.aacl-main.37} {{R}ec{I}n{D}ial: A unified framework for conversational recommendation with pretrained language models}.
\newblock In \emph{Proceedings of the 2nd Conference of the Asia-Pacific Chapter of the Association for Computational Linguistics and the 12th International Joint Conference on Natural Language Processing (Volume 1: Long Papers)}, pages 489--500, Online only. Association for Computational Linguistics.

\bibitem[{Wang et~al.(2022{\natexlab{b}})Wang, Su, and Chen}]{barcor}
Ting-Chun Wang, Shang-Yu Su, and Yun-Nung Chen. 2022{\natexlab{b}}.
\newblock \href {https://arxiv.org/pdf/2203.14257.pdf} {Barcor: Towards a unified framework for conversational recommendation systems}.
\newblock \emph{ArXiv}, abs/2203.14257.

\bibitem[{Wang et~al.(2022{\natexlab{c}})Wang, Zhou, Wen, and Zhao}]{unicrs}
Xiaolei Wang, Kun Zhou, Ji-Rong Wen, and Wayne~Xin Zhao. 2022{\natexlab{c}}.
\newblock \href {https://doi.org/10.1145/3534678.3539382} {Towards unified conversational recommender systems via knowledge-enhanced prompt learning}.
\newblock In \emph{Proceedings of the 28th ACM SIGKDD Conference on Knowledge Discovery and Data Mining}, page 1929–1937. Association for Computing Machinery.

\bibitem[{Wei et~al.(2021)Wei, Bosma, Zhao, Guu, Yu, Lester, Du, Dai, and Le}]{flan}
Jason Wei, Maarten Bosma, Vincent~Y Zhao, Kelvin Guu, Adams~Wei Yu, Brian Lester, Nan Du, Andrew~M Dai, and Quoc~V Le. 2021.
\newblock Finetuned language models are zero-shot learners.
\newblock \emph{arXiv preprint arXiv:2109.01652}.

\bibitem[{Wolf et~al.(2020)Wolf, Debut, Sanh, Chaumond, Delangue, Moi, Cistac, Rault, Louf, Funtowicz, Davison, Shleifer, von Platen, Ma, Jernite, Plu, Xu, Le~Scao, Gugger, Drame, Lhoest, and Rush}]{transformers}
Thomas Wolf, Lysandre Debut, Victor Sanh, Julien Chaumond, Clement Delangue, Anthony Moi, Pierric Cistac, Tim Rault, Remi Louf, Morgan Funtowicz, Joe Davison, Sam Shleifer, Patrick von Platen, Clara Ma, Yacine Jernite, Julien Plu, Canwen Xu, Teven Le~Scao, Sylvain Gugger, Mariama Drame, Quentin Lhoest, and Alexander Rush. 2020.
\newblock \href {https://doi.org/10.18653/v1/2020.emnlp-demos.6} {Transformers: State-of-the-art natural language processing}.
\newblock In \emph{Proceedings of the 2020 Conference on Empirical Methods in Natural Language Processing: System Demonstrations}, pages 38--45, Online. Association for Computational Linguistics.

\bibitem[{Wu et~al.(2023)Wu, Zheng, Qiu, Wang, Gu, Shen, Qin, Zhu, Zhu, Liu, Xiong, and Chen}]{wu2023aso}
Likang Wu, Zhilan Zheng, Zhaopeng Qiu, Hao Wang, Hongchao Gu, Tingjia Shen, Chuan Qin, Chen Zhu, Hengshu Zhu, Qi~Liu, Hui Xiong, and Enhong Chen. 2023.
\newblock \href {https://arxiv.org/pdf/2305.19860.pdf} {A survey on large language models for recommendation}.
\newblock \emph{ArXiv}, abs/2305.19860.

\bibitem[{Yang et~al.(2022)Yang, Han, Li, Zuo, and Yu}]{mese}
Bowen Yang, Cong Han, Yu~Li, Lei Zuo, and Zhou Yu. 2022.
\newblock \href {https://doi.org/10.18653/v1/2022.findings-naacl.4} {Improving conversational recommendation systems{'} quality with context-aware item meta-information}.
\newblock In \emph{Findings of the Association for Computational Linguistics: NAACL 2022}, pages 38--48, Seattle, United States. Association for Computational Linguistics.

\bibitem[{Zhang et~al.(2023{\natexlab{a}})Zhang, Han, Zhou, Hu, Yan, Lu, Li, Gao, and Qiao}]{llamaadapter}
Renrui Zhang, Jiaming Han, Aojun Zhou, Xiangfei Hu, Shilin Yan, Pan Lu, Hongsheng Li, Peng Gao, and Yu~Qiao. 2023{\natexlab{a}}.
\newblock \href {https://arxiv.org/pdf/2303.16199.pdf} {Llama-adapter: Efficient fine-tuning of language models with zero-init attention}.
\newblock \emph{ArXiv}, abs/2303.16199.

\bibitem[{Zhang et~al.(2019)Zhang, Kishore, Wu, Weinberger, and Artzi}]{zhang2019bertscore}
Tianyi Zhang, Varsha Kishore, Felix Wu, Kilian~Q Weinberger, and Yoav Artzi. 2019.
\newblock Bertscore: Evaluating text generation with bert.
\newblock \emph{arXiv preprint arXiv:1904.09675}.

\bibitem[{Zhang et~al.(2022)Zhang, Liu, Li, Zhong, Zhang, Wang, and Miao}]{kecrs}
Tong Zhang, Yong Liu, Boyang Li, Peixiang Zhong, Chen Zhang, Hao Wang, and Chunyan Miao. 2022.
\newblock \href {https://doi.org/10.18653/v1/2022.nlp4convai-1.17} {Toward knowledge-enriched conversational recommendation systems}.
\newblock In \emph{Proceedings of the 4th Workshop on NLP for Conversational AI}, pages 212--217, Dublin, Ireland. Association for Computational Linguistics.

\bibitem[{Zhang et~al.(2023{\natexlab{b}})Zhang, Xin, Li, Liu, Ren, Chen, Ma, and Ren}]{vricr}
Xiaoyu Zhang, Xin Xin, Dongdong Li, Wenxuan Liu, Pengjie Ren, Zhumin Chen, Jun Ma, and Zhaochun Ren. 2023{\natexlab{b}}.
\newblock \href {https://doi.org/10.1145/3539597.3570426} {Variational reasoning over incomplete knowledge graphs for conversational recommendation}.
\newblock In \emph{Proceedings of the Sixteenth {ACM} International Conference on Web Search and Data Mining}. {ACM}.

\bibitem[{Zhang et~al.(2020)Zhang, Sun, Galley, Chen, Brockett, Gao, Gao, Liu, and Dolan}]{dialogpt}
Yizhe Zhang, Siqi Sun, Michel Galley, Yen-Chun Chen, Chris Brockett, Xiang Gao, Jianfeng Gao, Jingjing Liu, and Bill Dolan. 2020.
\newblock \href {https://doi.org/10.18653/v1/2020.acl-demos.30} {{DIALOGPT} : Large-scale generative pre-training for conversational response generation}.
\newblock In \emph{Proceedings of the 58th Annual Meeting of the Association for Computational Linguistics: System Demonstrations}, pages 270--278, Online. Association for Computational Linguistics.

\bibitem[{Zhou et~al.(2021)Zhou, Wang, He, and Hou}]{crfr}
Jinfeng Zhou, Bo~Wang, Ruifang He, and Yuexian Hou. 2021.
\newblock \href {https://doi.org/10.18653/v1/2021.emnlp-main.355} {{CRFR}: Improving conversational recommender systems via flexible fragments reasoning on knowledge graphs}.
\newblock In \emph{Proceedings of the 2021 Conference on Empirical Methods in Natural Language Processing}, pages 4324--4334, Online and Punta Cana, Dominican Republic. Association for Computational Linguistics.

\bibitem[{Zhou et~al.(2020{\natexlab{a}})Zhou, Zhao, Bian, Zhou, Wen, and Yu}]{kgsf}
Kun Zhou, Wayne~Xin Zhao, Shuqing Bian, Yuanhang Zhou, Ji-Rong Wen, and Jingsong Yu. 2020{\natexlab{a}}.
\newblock \href {https://doi.org/10.1145/3394486.3403143} {Improving conversational recommender systems via knowledge graph based semantic fusion}.
\newblock In \emph{Proceedings of the 26th ACM SIGKDD International Conference on Knowledge Discovery and Data Mining}, page 1006–1014. Association for Computing Machinery.

\bibitem[{Zhou et~al.(2020{\natexlab{b}})Zhou, Zhou, Zhao, Wang, and Wen}]{tgredial}
Kun Zhou, Yuanhang Zhou, Wayne~Xin Zhao, Xiaoke Wang, and Ji-Rong Wen. 2020{\natexlab{b}}.
\newblock \href {https://doi.org/10.18653/v1/2020.coling-main.365} {Towards topic-guided conversational recommender system}.
\newblock In \emph{Proceedings of the 28th International Conference on Computational Linguistics}, pages 4128--4139, Barcelona, Spain (Online). International Committee on Computational Linguistics.

\bibitem[{Zhou et~al.(2022)Zhou, Zhou, Zhao, Wang, Jiang, and Hu}]{c2crs}
Yuanhang Zhou, Kun Zhou, Wayne~Xin Zhao, Cheng Wang, Peng Jiang, and He~Hu. 2022.
\newblock \href {https://doi.org/10.1145/3488560.3498514} {C²-crs: Coarse-to-fine contrastive learning for conversational recommender system}.
\newblock In \emph{Proceedings of the Fifteenth ACM International Conference on Web Search and Data Mining}, page 1488–1496. Association for Computing Machinery.

\bibitem[{Zou et~al.(2020)Zou, Chen, and Kanoulas}]{qrec}
Jie Zou, Yifan Chen, and Evangelos Kanoulas. 2020.
\newblock \href {https://doi.org/10.1145/3397271.3401180} {Towards question-based recommender systems}.
\newblock In \emph{Proceedings of the 43rd International ACM SIGIR Conference on Research and Development in Information Retrieval}, page 881–890. Association for Computing Machinery.

\end{thebibliography}
\bibliographystyle{acl_natbib}

\appendix

\section{System Outputs}
\label{sec:a}

We show an example from PECRS-medium on the INSPIRED dataset, in the same format as \Cref{fig:1}.

\begin{figure}[h]
    \includegraphics[width=\columnwidth]{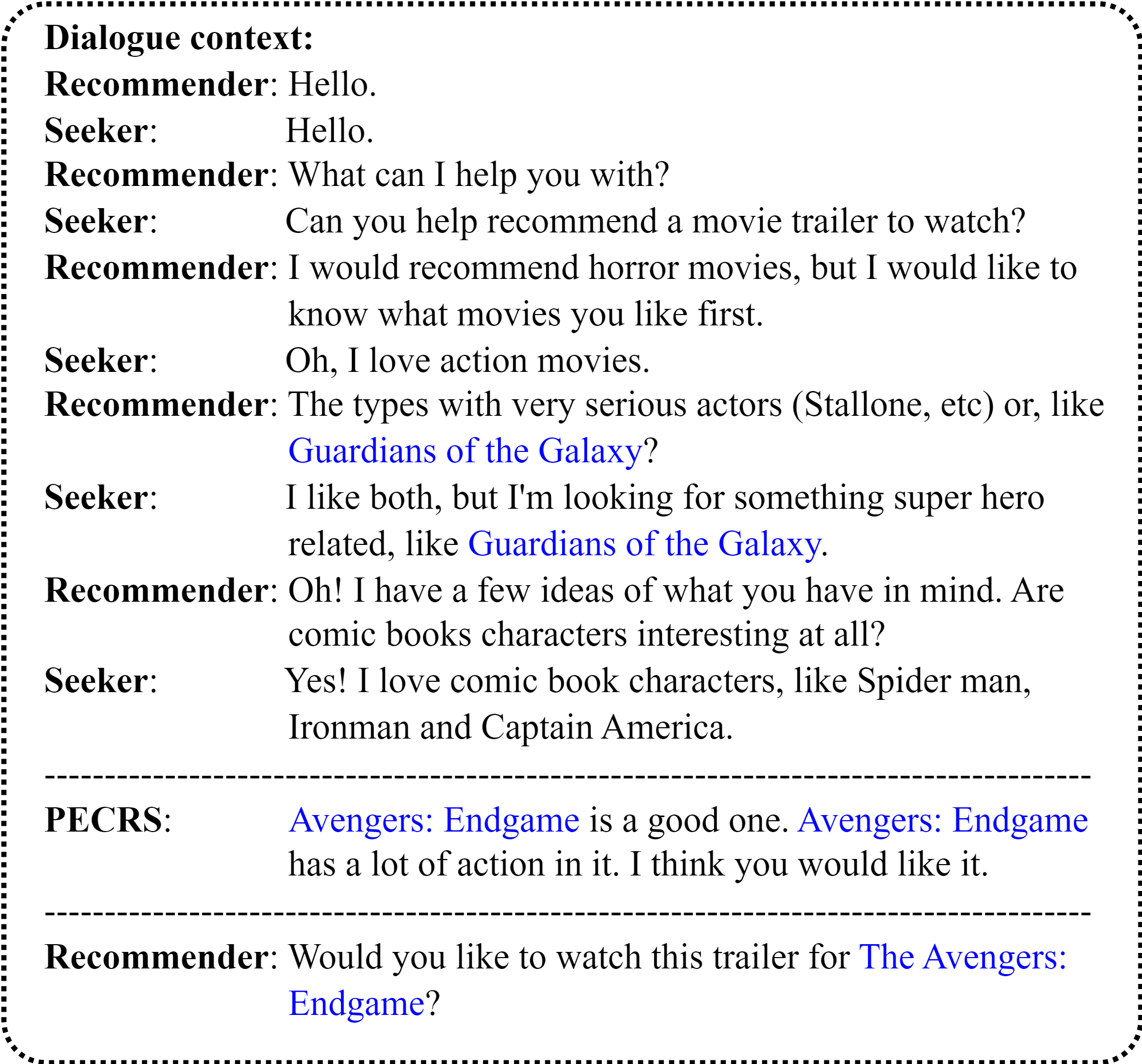}
    \caption{An example of dialogue from INSPIRED~\cite{inspired}, where \textcolor{blue}{blue} color denotes the movie items.}
    \label{fig:a}
\end{figure}

\section{Genre Analysis}
\label{sec:b}
In this section, we conduct a fine-grained analysis of PECRS top-1 recommendation. We investigate how the model performs on several types of items. To categorize items, we use the first genre tag in the Genre(s) field in the items metadata, yielding a partition of the movies set into 25 unique genres for ReDial, 22 genres for INSPIRED. We report the fraction of data points where the model outputs a top-1 movie of the correct genre per genre on ReDial and INSPIRED in \Cref{tab:b_1} and \Cref{tab:b_2}, respectively.

As we can see, there is wide variance in genres accuracy. Among wrong movie predictions, PECRS-medium outputs the correct genre 41.20\% times on ReDial and 30.04\% on INSPIRED. Random performance would yield 16.26\% and 19.39\% accuracy, respectively. The performance is much higher on highly represented genres such as \emph{Comedy}, \emph{Action}, or \emph{Horror}, where it can surpass a ratio of correctly predicted genre of 50\%, but quickly falls to 0 for rare genres such as \emph{Romance}. Future work may focus on better handling the long tail distribution in items variety, for instance through data augmentation techniques crafted for rare genres movies.

\begin{table}[]
\setlength\extrarowheight{-3pt}
\resizebox{\columnwidth}{!}{  
\begin{tabular}{lccc}

\toprule 

\textbf{Genre} 
& \textbf{Items (\%)} 
& \textbf{\begin{tabular}[c]{@{}c@{}}Test set\\ Recommendation (\%)\end{tabular}}
& \textbf{\begin{tabular}[c]{@{}c@{}}Correctly\\ Predicted (\%)\end{tabular}} \\

\midrule 

Comedy         & 24.48 & 23.74 & 46.37 \\
Action         & 21.88 & 30.67 & 57.65 \\
Drama          & 17.89 & 13.74 & 32.93 \\
Adventure      & 6.18 & 6.02 & 29.72 \\
Horror         & 5.82 & 7.94 & 46.95 \\
Crime          & 5.66 & 4.50 & 20.56 \\
Animation      & 5.40 & 6.71 & 62.38 \\
Biography      & 4.11 & 2.50 & 12.61 \\
Documentary    & 3.27 & 0.76 & 22.22 \\
Fantasy        & 1.00 & 0.61 & 6.90 \\
Thriller       & 0.67 & 0.46 & 31.82 \\
Family         & 0.62 & 0.38 & 0.00 \\
Mystery        & 0.47 & 0.57 & 7.41 \\
Romance        & 0.46 & 0.04 & 0.00 \\
TV             & 0.43 & 0.08 & 0.00 \\
Music          & 0.26 & 0.20 & 0.00 \\
Western        & 0.25 & 0.04 & 0.00 \\
Science        & 0.23 & 0.13 & 0.00 \\
Short          & 0.23 & 0.11 & 0.00 \\
War            & 0.21 & 0.11 & 0.00 \\
Sci-fi         & 0.20 & 0.06 & 0.00 \\
History        & 0.11 & 0.00 & \_ \\
Musical        & 0.10 & 0.23 & 9.09 \\
Film-noir      & 0.05 & 0.08 & 0.00 \\
Adult          & 0.02 & 0.02 & 0.00 \\

\bottomrule 

\end{tabular}
}
\caption{Accuracy w.r.t genre prediction on ReDial test set broken down by movie genre.}
\label{tab:b_1}
\end{table}

\begin{table}[h]
\setlength\extrarowheight{-3pt}
\resizebox{\columnwidth}{!}{  
\begin{tabular}{lccc}

\toprule 

\textbf{Genre} 
& \textbf{Items (\%)} 
& \textbf{\begin{tabular}[c]{@{}c@{}}Test set\\ Recommendation (\%)\end{tabular}}
& \textbf{\begin{tabular}[c]{@{}c@{}}Correctly\\ Predicted (\%)\end{tabular}} \\

\midrule 

Action         & 24.01 & 36.20 & 50.50 \\
Comedy         & 22.66 & 17.92 & 52.00 \\
Drama          & 17.67 & 13.98 & 10.26 \\ 
Horror         & 7.45 & 9.68 & 14.81 \\
Adventure      & 4.86 & 2.15 & 66.67 \\ 
Animation      & 4.86 & 4.66 & 7.69 \\ 
Crime          & 4.86 & 6.09 & 23.53 \\
Biography      & 4.50 & 2.15 & 0.00 \\
Documentary    & 3.20 & 1.79 & 0.00 \\
Thriller       & 0.92 & 0.36 & 0.00\\
Fantasy        & 0.86 & 0.36 & 0.00\\ 
Romance        & 0.80 & 0.36 & 0.00 \\ 
Mystery        & 0.62 & 0.00 & \_ \\ 
TV             & 0.37 & 0.00 & \_ \\ 
Short          & 0.37 & 0.00 & \_ \\ 
Science        & 0.31 & 0.72 & 0.00 \\ 
Music          & 0.25 & 0.00 & \_ \\ 
Sci-fi         & 0.25 & 0.36 & 0.00 \\ 
War            & 0.12 & 0.00 & \_ \\ 
Western        & 0.12 & 0.00 & \_ \\ 
Musical        & 0.06 & 0.00 & \_ \\
Reality-TV     & 0.06 & 0.00 & \_ \\ 

\bottomrule 

\end{tabular}
}
\caption{Accuracy w.r.t genre prediction on INSPIRED test set broken down by movie genre.}
\label{tab:b_2}
\end{table}

\section{Packages}
\label{sec:c}
Our framework was implemented in Python 3.8.0. We used the following Python package versions to conduct all experiments:
\begin{itemize}
    \item numpy 1.24.3
    \item torch 1.9.1
    \item transformers 4.33.2
    \item rouge-score 0.1.2
    \item nltk 3.8.1
    \item peft 0.1.0
    \item spacy 3.6.0
\end{itemize}

All packages and datasets used are freely available and open-source, and were used for research purpose only. We refer to the specific papers for more details on the use of each dataset.



\end{document}